\documentclass{article}
\usepackage{kr}

\usepackage{times}
\usepackage{soul}
\usepackage{url}
\usepackage[hidelinks]{hyperref}
\usepackage[utf8]{inputenc}
\usepackage[small]{caption}
\usepackage{graphicx}
\usepackage{subfigure}
\usepackage{amsmath}
\usepackage{amssymb}
\usepackage{amsthm}
\usepackage{booktabs}
\usepackage{algorithm}
    \usepackage[noend]{algorithmic}
\usepackage{todonotes}
\newtheorem{example}{Example}

\newtheorem{notation}{Notation}
\newtheorem{proposition}{Proposition}
\newtheorem{definition}{Definition}
\newtheorem{corollary}{Corollary}
\newtheorem{metric}{Metric}

\title{CE-QArg: Counterfactual Explanations for\\Quantitative Bipolar Argumentation Frameworks}

\author{%
Xiang Yin$^1$\and
Nico Potyka$^2$\and
Francesca Toni$^1$\\
\affiliations
$^1$Department of Computing, Imperial College London, UK\\
$^2$School of Computer Science and Informatics, Cardiff University, UK\\
\emails
\{xy620, ft\}@imperial.ac.uk,
potykan@cardiff.ac.uk
}

\begin{document}

\maketitle

\begin{abstract}
There is a growing interest in understanding arguments' strength in Quantitative Bipolar Argumentation Frameworks (QBAFs).
Most existing studies focus on attribution-based methods that explain an argument's strength by assigning importance scores to other arguments but fail to explain how to change the current strength to a desired one.
To solve this issue, we introduce \emph{counterfactual explanations} for QBAFs.
We discuss problem variants and
propose an iterative algorithm named \emph{Counterfactual Explanations for Quantitative bipolar Argumentation frameworks (CE-QArg)}. CE-QArg can identify \emph{valid} and \emph{cost-effective} counterfactual explanations based on two core modules, \emph{polarity} and \emph{priority}, which help determine the updating direction and magnitude for each argument, respectively.
We discuss some formal properties of our counterfactual explanations and empirically evaluate CE-QArg
on randomly generated QBAFs.
\end{abstract}

\section{Introduction}
\label{sec_intro}
Explainable AI (XAI) aims to enhance the transparency and trustworthiness of AI models by providing explanations for their decision-making process~\cite{adadi2018peeking}, which is crucial  in high-stakes decision-making domains such as healthcare, finance, and judiciary.
Recently, explaining the reasoning process of Quantitative Bipolar Argumentation Frameworks (QBAFs)~\cite{baroni2015automatic} has received increasing attention~\cite{kampik2022explaining,AAE_ECAI,kampik2024contribution}.
QBAFs consist of \emph{arguments}, binary relations (of \emph{support} and \emph{attack}), and a \emph{base score function} that ascribes initial strengths to each argument. QBAFs semantics typically 
determine each argument's \emph{(final) strength} based on the strength of its attackers and supporters (e.g. see \cite{leite2011social,baroni2015automatic,amgoud2018evaluation}), which allows quantitative reasoning among contradictory information~\cite{vcyras2021argumentative,potyka2021interpreting,ayoobi2023sparx,potyka2023explaining}.
To explain arguments' final strength, often attribution-based methods
are applied (e.g. see \cite{kampik2022explaining,kampik2024contribution,AAE_ECAI}). These methods assign ``importance scores" to arguments, showing how much they contribute to the final strength of arguments of interest. 

\begin{figure}[t]
    \centering
    \includegraphics[width=0.8\columnwidth]{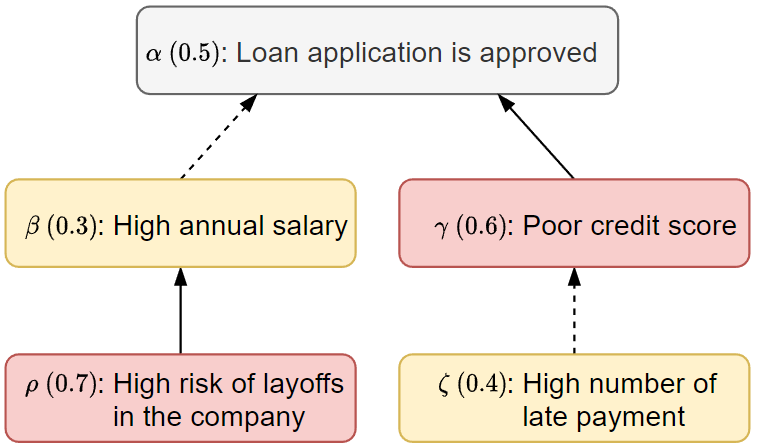}
    \caption{An example QBAF for loan application. 
    (
    Solid and dashed edges indicate \emph{attack} and \emph{support}, respectively; the numbers in brackets are the arguments' base scores).
    }
    \label{fig_loan}
\end{figure}

To illustrate the idea, Figure~\ref{fig_loan} shows
a QBAF to decide whether a person's loan application
will be approved.
This QBAF has a hierarchical structure, where at the top level is the \emph{topic argument} $\alpha$, determining whether the loan will be approved. 
Two other arguments influence $\alpha$:
(i) $\beta$ (high annual salary) supports $\alpha$, but it is also attacked by $\rho$ (high risk of layoffs);
(ii) $\gamma$ (poor credit score) attacks $\alpha$, and it is supported by $\zeta$ (high number of late payments).
The base score of $\alpha$ is initially set to 0.5, while the base scores for $\beta$, $\gamma$, $\rho$, and $\zeta$ are 0.3, 0.6, 0.7, and 0.4, respectively.
We apply the DF-QuAD semantics~\cite{rago2016discontinuity} (denoted as $\sigma_{DF}$) to evaluate this QBAF: the loan will be approved if $\sigma_{DF}(\alpha) \geq 0.5$, and rejected otherwise. 
In this scenario, the bank rejects the applicant because the final strength is lower than the desired strength ($\sigma_{DF}(\alpha)=0.165<0.5$). 
To explain such an outcome, 
for instance, the Shapley-based attribution method~\cite{kampik2024contribution}, finds that $\beta$ has a positive importance score of $0.0975$ wrt. $\alpha$, while $\gamma$, $\zeta$, and $\rho$ all have negative importance scores of $-0.34$, $-0.0525$ and $-0.04$, respectively\footnote{See Section~\ref{sec_preliminary} and \url{https://arxiv.org/abs/2407.08497} for more details of the computation of strengths and importance scores, respectively.}. Since the negative scores outweigh the positive one, the final strength is lower than the desired strength, resulting in an unsuccessful application.

While attribution explanations are intuitive, in this example and more generally,
they fail to offer guidance on how to modify the topic argument's strength, e.g., in the example, to improve one's chance of getting approved. 
In contrast, \emph{counterfactual explanations} \cite{wachter2017counterfactual} explain how an AI model's output would change if one alters the inputs. These explanations are comprehensible because they elicit causal reasoning and thinking in humans~\cite{byrne2019counterfactuals,verma2020counterfactual}. 

In this work, we introduce counterfactual explanations to the QBAF setting to compensate for the limitations of attribution explanations.
Here, counterfactually explaining a QBAF means
to identify a base score function that could lead to a specified final strength of a topic argument under a given gradual semantics.
For example, in Figure~\ref{fig_loan}, if the applicant had a different base score function that could result in a desired final strength of $\alpha$, then the loan would have been approved.

The main contributions of our paper are:
\begin{itemize}
    \item We formally define three counterfactual problems for QBAFs (Section~\ref{sec_theory}).
    \item We propose formal properties guiding the design of explanation methods to solve  counterfactual problems for QBAFs and propose an iterative algorithm (CE-QArg) to generate valid, cost-effective counterfactuals\footnote{We sometimes refer to counterfactual explanations simply as counterfactuals.} (Section~\ref{sec_algo}).
    \item We propose formal properties for counterfactuals (Section~\ref{sec_properties}).
    \item We empirically show the high effectiveness, scalability, and robustness of CE-QArg (Section~\ref{sec_evaluation}).
\end{itemize}
The proofs of all technical results are in
\url{https://arxiv.org/abs/2407.08497}.
The code is available at \url{https://github.com/XiangYin2021/CE-QArg}.

\section{Related Work}
\label{sec_relatedwork}
The literature on QBAF explanations primarily focuses on attribution-based explanations.
These methods aim to explain the reasoning outcome for (final strength of) a particular argument (
referred to as the topic argument) in a QBAF by assigning importance scores to arguments.
There are various ways to define these importance scores, including removal-based methods~\cite{kampik2024contribution}
that measure the impact of removing an 
argument;
gradient-based methods~\cite{AAE_ECAI}
that measure 
sensitivity with respect 
to an argument's base score; 
and Shapley-based methods~\cite{kampik2024contribution} that distribute the overall impact among all arguments
using tools from game theory.
Shapley-based methods can also be used to attribute importance scores to edges (attacks and supports) rather than arguments in a QBAF 
to explain their impact~\cite{amgoud2017measuring,YIN_RAE_IJCAI}.
While attribution-based explanations are helpful and intuitive in explaining 
outcomes (final strength), they cannot guide improving them 
by altering the inputs (
QBAFs). 
We focus instead on counterfactual 
explanations.

Counterfactual explanations are typically used in XAI \cite{wachter2017counterfactual} and \emph{contestable} AI \cite{alfrink2023contestable,leofante2024contestable} to indicate paths towards ``algorithmic recourse''. 
Counterfactual explanations for QBAFs have not been well-studied.
A recent study by \cite{kampik2024change} focuses on explaining why the strengths' partial order of two topic arguments swap after updating QBAFs (e.g. by adding/removing arguments/edges or changing the base scores of arguments), seeing 
counterfactual explanations as argument sets whose elements, when updated, cause this swap
. 
For example, suppose there are two topic arguments $\alpha$ and $\beta$ in a QBAF and the final strength of $\alpha$ is smaller than that of $\beta$.
If a supporter $\gamma$ for $\alpha$ were added
, resulting in $\alpha$'s strength becoming larger than $\beta$'s, then the set $\{\gamma\}$ is a possible counterfactual explanation for the strength swap between $\alpha$ and $\beta$.
Instead, we focus on explaining arguments' strength rather than partial orders 
between arguments' strengths. Additionally, we focus on counterfactuals as
 base score functions in structure-fixed QBAFs rather than  as argument sets in structure-changeable QBAFs.
Another work \cite{OrenYVB22}, implicitly relates to counterfactual explanations for QBAFs under gradual semantics, by studying the inverse problem 
of identifying base score functions that can lead to a desired ranking of strengths for all arguments. Differently, we focus on reaching a desired strength for topic arguments instead of a desired ranking of all arguments.

It is also worth mentioning \cite{sakama2014counterfactual}, which initially introduced and investigated counterfactual problems in the argumentation area. This work studies what would happen if an initially accepted (rejected) argument were rejected (accepted) (e.g. by adding a new attacker or removing all the attackers 
towards an argument) and how to explain the corresponding acceptance change of 
arguments. 
Differently from our method, this work focuses on abstract argumentation 
under complete labellings~\cite{dung1995acceptability} and 
on the consequences of changes rather than their causes as we do.

\section{Preliminaries}
\label{sec_preliminary}
Formally, a QBAF can be defined as follows.
\begin{definition}
\label{def_qbaf}
A \emph{Quantitative Bipolar Argumentation Framework (QBAF)} is a quadruple $\mathcal{Q}=\langle \mathcal{A}, \mathcal{R}^{-}, \mathcal{R}^{+}, \tau \rangle$ consisting of a finite set of \emph{arguments} $\mathcal{A}$, binary relations of \emph{attack} $\mathcal{R}^{-} \subseteq \mathcal{A} \times \mathcal{A}$ and \emph{support} $\mathcal{R}^{+} \subseteq \mathcal{A} \times \mathcal{A}$ $(\mathcal{R}^{-} \cap \mathcal{R}^{+} = \emptyset)$ and a \emph{base score function} $\tau:\mathcal{A} \rightarrow [0,1]$.
\end{definition}

The base score function in QBAFs ascribes initial strengths (base scores) to arguments therein.
QBAFs may be represented graphically (as in Figure~\ref{fig_loan}) using nodes to represent arguments and their base scores and edges to show the relations among arguments. Then QBAFs are said to be \emph{(a)cyclic} if the graphs representing them are (a)cyclic.

\begin{definition}
\label{def_semantics}
A \emph{gradual semantics} $\sigma$ is a function 
that evaluates a QBAF $\mathcal{Q}=\langle \mathcal{A}, \mathcal{R}^{-}, \mathcal{R}^{+}, \tau \rangle$ by ascribing  values $\sigma(\alpha) \in [0,1]$ to every $\alpha \in \mathcal{A}$ as their \emph{
strength}.
\end{definition}

Different (gradual) semantics typically ascribe different strengths to arguments
.
Most semantics define the strength of an argument through an iterative procedure involving two functions: first, an \emph{aggregation function} aggregates the strength of the argument's attackers and supporters; then, an \emph{influence function} combines the aggregation values with the argument's base score to determine its 
strength.
Gradual semantics guarantee convergence for acyclic QBAFs~\cite{Potyka19}. For cyclic QBAFs,
the strength values may not converge \cite{mossakowski2018modular},
but when they do, they converge quickly in practice~\cite{Potyka18}.
Our focus in this paper is on cases where convergence
occurs, as we aim to explain the strength, which is only possible when the strength is defined. 
Thus, we will assume in the remainder that 
gradual semantics are \emph{well-defined}, as follows. 

\begin{definition}
A gradual semantics $\sigma$ is \emph{well-defined}
for a QBAF $\mathcal{Q}=\langle \mathcal{A}, \mathcal{R}^{-}, \mathcal{R}^{+}, \tau \rangle$ iff $\sigma(\alpha)$ exists for every $\alpha \in \mathcal{A}$.
\end{definition}


\begin{notation}
\label{def_cf_strength}
In the remainder, unless specified otherwise, we 
use $\mathcal{Q}\!=\!\langle \mathcal{A}, \mathcal{R}^{-}, \mathcal{R}^{+}, \tau \rangle$ and $\mathcal{Q}_{\tau'} \!=\! \langle \mathcal{A}, \mathcal{R}^{-}, \mathcal{R}^{+}, \tau' \rangle$ to indicate two 
QBAFs with different base score functions only, and use $\sigma$ for any (well-defined) gradual semantics.
For $\alpha\in \mathcal{A}$, we let 
$\sigma_{\tau'}(\alpha)$ denote 
$\sigma(\alpha)$ in $\mathcal{Q}_{\tau'}$.
\end{notation}
Concretely, we will use the Quadratic Energy (QE)~\cite{Potyka18}, Restricted Euler-based (REB)~\cite{amgoud2018evaluation}, and Discontinuity-Free Quantitative Argumentation Debates (DF-QuAD)~\cite{rago2016discontinuity}. To aid understanding, we will show the definition and an example of the DF-QuAD semantics and recap the other two in \url{https://arxiv.org/abs/2407.08497}.

In DF-QuAD, for any argument $\alpha \in \mathcal{A}$, $\sigma^{DF}(\alpha)$ is defined as follows:

Aggregation function:
$$
E_\alpha = \prod_{\left \{ \beta \in \mathcal{A} \mid (\beta,\alpha) \in \mathcal{R^{-}} \right \} }\!\!\!\!\!\!\!\!\!\!\!\!\!\!(1-\sigma^{DF}(\beta)) - \prod_{\left \{ \beta \in \mathcal{A} \mid (\beta,\alpha) \in \mathcal{R^{+}} \right \} }\!\!\!\!\!\!\!\!\!\!\!\!\!\!(1-\sigma^{DF}(\beta)).
$$

Influence function:
$$
\sigma^{DF}(\alpha)= 
\begin{cases}
    \tau(\alpha)-\tau(\alpha) \cdot \left|E_\alpha\right| & if\ E_\alpha \leq 0;\\
    \tau(\alpha)+(1-\tau(\alpha))\cdot E_\alpha & if\ E_\alpha > 0.\\
\end{cases}
$$
An example of applying DF-QuAD is as follows.
\begin{example}
Considering the QBAF in Figure~\ref{fig_loan}, 
where $\tau(\alpha)=0.5$, $\tau(\beta)=0.3$, $\tau(\gamma)=0.6$, $\tau(\rho)=0.7$, and $\tau(\zeta)=0.4$.
According to the aggregation and influence function of the DF-QuAD semantics\footnote{We follow the convention that the product of an empty set is $1$ (not $0$) because $1$ is the neutral element with respect to multiplication.}, we have $E_\alpha=0.24-0.91=-0.67$, $E_\beta=-0.7$, $E_\gamma=0.4$, $E_\rho=0$, $E_\zeta=0$,
$\sigma^{DF}(\alpha)=\tau(\alpha)-\tau(\alpha) \cdot \left|E_\alpha\right|=0.165$,
$\sigma^{DF}(\beta)=\tau(\beta)-\tau(\beta) \cdot \left|E_\beta\right|=0.09$,
$\sigma^{DF}(\gamma)=\tau(\gamma)+(1-\tau(\gamma)) \cdot E_\gamma=0.76$,
$\sigma^{DF}(\rho)=\tau(\rho)=0.7$,
$\sigma^{DF}(\zeta)=\tau(\zeta)=0.4$.
\end{example}


\section{Counterfactuals for QBAFs}
\label{sec_theory}

\begin{figure}[t]
    \centering
    \includegraphics[width=1.0\columnwidth]{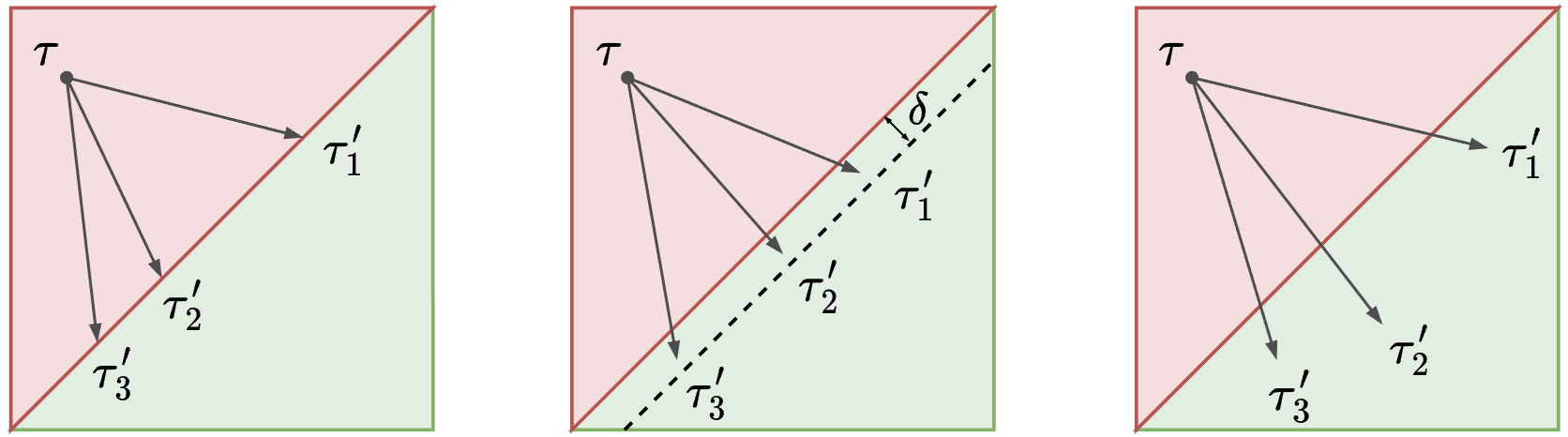}
    \caption{Illustration of strong (left), $\delta-$approximate (middle), and weak (right) counterfactual problems. The squares represent all possible base score functions, with $\tau$ the current base score function, and the red (above diagonal) and green (below diagonal) parts as undesirable and desirable alternatives, respectively.}
    \label{fig_intuition}
\end{figure}

We define three 
counterfactual problems (see Figure~\ref{fig_intuition}) and explore the existence of solutions  
thereto (i.e. \emph{counterfactuals}) 
and 
their relationships. 
Intuitively,
given a QBAF, a topic argument and a desired strength therefor, we 
see counterfactuals as changes to the 
base score function to obtain the desired strength for the topic argument. 


\subsection{Strong Counterfactual Problem}
\label{sec:strong}
\begin{definition}
\label{def_strong_cfx}
Given 
a topic argument $\alpha^* \in \mathcal{A}$
and a desired strength $s^*$ for $\alpha^*$ such that $\sigma(\alpha^*) \neq s^*$ in $\mathcal Q$,
the \emph{strong counterfactual problem} amounts to identifying  a base score function $\tau'\neq \tau$ such that 
$\sigma_{\tau'}(\alpha^*)=s^*$ (in $\mathcal{Q}_{\tau'}$).
\end{definition}

In Figure~\ref{fig_intuition} (left), 
$\tau$ is in the (undesired) red part 
and $\tau'_{1}, \tau'_{2},\tau'_{3}$ in the green part are 
possible 
counterfactuals (base score functions) 
as they exactly hit the desired strength.

A trivial solution to the strong counterfactual problem might be 
just setting the base score of the topic argument to the desired strength and the base score of all others to 0. 
\begin{definition}
\label{def_trivial_solution}
Given
a topic argument $\alpha^* \in \mathcal{A}$ 
and a desired strength $s^*$ for $\alpha^*$,
the \emph{trivial counterfactual
} is the base score function $\tau'\neq \tau$
such that $\tau'(\alpha^*)=s^*$ and $\tau'(\alpha)=0$ for all
$\alpha \in \mathcal{A} \setminus \{\alpha^*\}$
.
\end{definition}
The trivial counterfactual is a solution to the
strong counterfactual problem if the semantics 
satisfies the following stability property \cite{amgoud2018evaluation}.\footnote{Strictly speaking, the definition in \cite{amgoud2018evaluation} assumes 
$\mathcal{R}^{-} = \mathcal{R}^{+} = \emptyset$, but if the neutrality property \cite{amgoud2018evaluation} holds, the definitions are equivalent.}
\begin{definition}
A gradual semantics $\sigma$ satisfies \emph{s-stability} iff
for any $\alpha \in \mathcal{A}$,
whenever  $\sigma(\beta) = 0$ for all $(\beta, \alpha) \in \mathcal{R}^{-} \cup \mathcal{R}^{+}$,   
then $\sigma(\alpha) = \tau(\alpha)$.
\end{definition}

\begin{proposition}[Solution Existence]
\label{prop_solution_exixtence}
If 
$\sigma$ satisfies s-stability
and $\mathcal{Q}$ is acyclic,
then the trivial counterfactual is a solution to the strong counterfactual problem
.
\end{proposition}

\begin{proposition}
\label{prop_stability_satisfaction}
QE, REB, and DF-QuAD satisfy s-stability.
\end{proposition}

Since commonly considered semantics satisfy s-stability, a (trivial) solution to the strong counterfactual problem always exists  
in acyclic QBAFs.
For cyclic QBAFs, the trivial counterfactual is not a solution even in a very simple QBAF with two mutually supporting arguments
.

\subsection{$\delta-$Approximate Counterfactual Problem}
\label{sec:approx}
Ensuring that the final strength of a topic argument analytically equals 
the desired strength may be challenging due to the complexity of the 
definition of gradual semantics that may involve 
various linear and nonlinear transformations (e.g., QE, REB and DF-QuAD
). Thus, we consider the following relaxation
of the strong counterfactual problem.

\begin{definition}
\label{def_delta_cfx}
Given 
a 
constant $\delta\!>\!0$, 
a topic argument $\alpha^* \!\in\! \mathcal{A}$ 
and a desired strength $s^*$ for $\alpha^*$ such that $\sigma(\alpha^*) \neq s^*$ in $\mathcal Q$,
the \emph{$\delta-$approximate counterfactual problem} amounts to identifying  a base score function $\tau'\neq \tau$ 
such that 
\begin{itemize}
    \item if $\sigma(\alpha^*)\! <\! s^* $ in $\mathcal{Q}$, then  $s^*\! \leq\! \sigma_{\tau'}(\alpha^*) \!\leq\! s^*\!+\!\delta$ in $\mathcal{Q}_{\tau'}$ and
    \item if $\sigma(\alpha^*)\! >\! s^* $ in $\mathcal{Q}$, then  $s^*\!-\!\delta \!\leq \!\sigma_{\tau'}(\alpha^*)\!\leq \!s^*$ in $\mathcal{Q}_{\tau'}$.
\end{itemize}
\end{definition}

As an illustration, in Figure~\ref{fig_intuition} (middle), 
$\tau'_{1}, \tau'_{2}$ and $\tau'_{3}$ are all solutions to the $\delta-$approximate counterfactual problem as they all lie in the $\delta$-interval region of the desired strength.

\subsection{Weak Counterfactual Problem}
For practical applications, it is also worth further relaxing the $\delta-$approximate counterfactual problem, e.g. 
when QBAFs are 
used to solve binary classification problems~\cite{cocarascu2019extracting,kotonya2019gradual}, 
the topic argument's final strength does not necessarily need to equal or approximate the desired strength.
Concretely,
in an argumentative movie recommender system of~\cite{cocarascu2019extracting}, 
as long as a movie's rating (final strength) is higher than some threshold, the movie is considered 
of good quality. Thus, we next propose 
the following further relaxation of the strong counterfactual problem.




\begin{definition}
\label{def_weak_cfx}
Given 
a topic argument $\alpha^* \!\!\in\!\! \mathcal{A}$ in $\mathcal{Q}$
and a desired strength $s^*$ 
for $\alpha^*$ such that $\sigma(\alpha^*) \neq s^*$ in $\mathcal Q$, 
the \emph{weak counterfactual problem} amounts to identifying a base score function $\tau'\!\!\neq \!\!\tau$ 
such that
\begin{itemize}
    \item if $\sigma(\alpha^*) < s^* $ in $\mathcal{Q}$, then  $\sigma_{\tau'}(\alpha^*) \geq s^*$ in $\mathcal{Q}_{\tau'}$ and
    \item if $\sigma(\alpha^*) > s^* $ in $\mathcal{Q}$, then  $\sigma_{\tau'}(\alpha^*) \leq s^*$ in $\mathcal{Q}_{\tau'}$.
\end{itemize}
\end{definition}

As an illustration, in Figure~\ref{fig_intuition} (right), 
$\tau'_{1}, \tau'_{2}$ and $\tau'_{3}$ are all solution to the weak counterfactual problem as they all cross the threshold (red diagonal line). 




\subsection{Validity and Problem Relationships}
We define notions of validity for 
counterfactuals 
and then explore the relationships 
among these notions.

To 
be consistent 
with the literature of 
counterfactuals in XAI, if a base score function is a solution to a counterfactual problem, we say that this solution is a valid counterfactual.

\begin{definition}
\label{def_validity}
A \emph{valid counterfactual 
 for the strong/$\delta-$ap\-pro\-ximate/weak counterfactual problem} is a base score function $\tau'$ 
which is a solution to the strong/$\delta-$approximate/weak counterfactual problem, respectively.
\end{definition}

To illustrate, in Figure~\ref{fig_intuition}, all $\tau'_{1}, \tau'_{2}$ and $\tau'_{3}$ are valid counterfactuals as they are solutions to the (respective) problems.



Next, we study relationships among valid counterfactuals
.
\begin{proposition}[Problem Relationships]
\label{prop_rels}
\hfill\par
\begin{enumerate}
\item If a counterfactual is valid for the strong counterfactual problem, then it is also valid for the $\delta-$approximate and weak counterfactual problems.
\item If a counterfactual is valid for the $\delta-$approximate counterfactual problem, then it is also valid for the weak counterfactual problem.
\end{enumerate}
\end{proposition}

\begin{corollary}
If 
$\sigma$ satisfies s-stability
and $\mathcal{Q}$ is acyclic,
then
the trivial counterfactual 
is 
valid for the strong/$\delta-$approximate/weak counterfactual problem
.
\end{corollary}

However, non-trivial valid counterfactuals do not always exist when not allowing directly setting the base score of the topic argument to the desired strength. 
For instance, when a topic argument is not connected with any other arguments in the QBAF, then there is no non-trivial counterfactual.

\begin{proposition}[Uniqueness]
\label{prop_uniqueness}
There is a unique valid counter\-factual 
(for the strong/$\delta-$approximate/weak counter\-factual problem) iff the trivial counterfactual is the only valid counterfactual 
(for the 
respective problem).
\end{proposition}



We leave to future study the identification of special classes of QBAFs for which non-trivial counterfactual explanations can be always guaranteed to exist.

\section{Cost-Effective  Counterfactuals}
\label{sec_algo}


In this section we turn to 
computational challenges: 
we aim to design an algorithm that can return not only \textbf{\emph{valid}} but also \textbf{\emph{cost-effective}} counterfactuals for the $\delta-$approximate counterfactual problem. We focus on this problem because, as discussed in Section~\ref{sec:approx},  
it may be 
unrealistic to aim at exactly matching the desired strength and solve the strong counterfactual problem and,  by  Proposition~\ref{prop_rels},
once a valid $\delta-$approximate counterfactual is returned, it is also valid for the weak counterfactual problem. 
We interpret cost as distance: 
the shorter the distance, the lower the cost of a counterfactual.
This is in line with literature on counterfactual explanations in XAI~\cite{wachter2017counterfactual}.
To illustrate, in Figure~\ref{fig_intuition} (middle), while  $\tau'_{1}, \tau'_{2}$ and $\tau'_{3}$ are all valid $\delta-$approximate, their distance to
the original base score varies, using the following notion of distance. 

\begin{definition}
\label{def_distance_Lp}
The \emph{$L_p$-Norm Distance} between $\tau$ and $\tau'$ is:
$$d_p(\tau,\tau')=\left(\sum_{\alpha \in \mathcal{A}}||\tau(\alpha)-\tau'(\alpha)||^p\right)^{\frac{1}{p}}.$$
\end{definition}

Minimizing this distance is difficult because there is no closed expression for a topic argument's final strength
in general (cyclic) graphs.
Here, we will therefore propose an approximate algorithm and evaluate it empirically in Section~\ref{sec_evaluation}.
Before giving the algorithm (in Section~\ref{subsec:algo}), we will
present properties driving its design (Section~\ref{subsec:des} below).

\subsection{Algorithm Design Properties}
\label{subsec:des}

We will design an iterative algorithm that incrementally adapts the weight to find a close counterfactual.
To do so, we need to determine an updating \textbf{\emph{direction}} and \textbf{\emph{magnitude}}, which means 
deciding how much each argument's base score should be increased or decreased. 
%
To determine the 
former, we partition arguments based on their \emph{polarity}.
We first define \emph{paths} and \emph{connectivity} between arguments.

\begin{definition}
\label{def:path}
For any $\alpha, \beta \!\in\! \mathcal{A}$,
we let $p_{\alpha \mapsto \beta}\!=\!\langle(\gamma_0,\gamma_1),$ $ (\gamma_1,\gamma_2),\cdots,(\gamma_{n-1},\gamma_{n})\rangle(n \geq 1)$ denote a \emph{path} from $\alpha$ to $\beta$, 
where $\alpha\!=\!\gamma_0$, $\beta\!=\!\gamma_n$, 
$\gamma_{i} \in \mathcal{A} (1 \leq i \leq n)$
and $(\gamma_{i-1},\gamma_{i}) 
\in 
\mathcal{R}^{-} \cup \mathcal{R}^{+}$.
\end{definition}

\begin{notation}
We let 
$\left| p_{\alpha \mapsto \beta} \right|$ denote the length of path $p_{\alpha \mapsto \beta}$. 
We let $P_{\alpha \mapsto \beta}$ and $|P_{\beta \mapsto \alpha}|$ denote the set of all paths from $\alpha$ to $\beta$, and the number of paths in $P_{\alpha \mapsto \beta}$, respectively.
\end{notation}

We next distinguish three types of connectivity based on the number of paths from one argument to another.
\begin{definition}
\label{def_connectivity}
For any $\alpha,\beta \in \mathcal{A}$: 
\begin{itemize}
    \item $\beta$ is \emph{disconnected} from $\alpha$ iff 
    $P_{\beta \mapsto \alpha}=\emptyset$;
    \item $\beta$ is \emph{single-path connected} to $\alpha$ iff 
    $|P_{\beta \mapsto \alpha}|=1$;
    \item $\beta$ is \emph{multi-path connected} to $\alpha$ iff 
    $|P_{\beta \mapsto \alpha}|>1$.
\end{itemize}
\end{definition}

\begin{figure}[t]
    \centering
    \includegraphics[width=0.55\columnwidth]{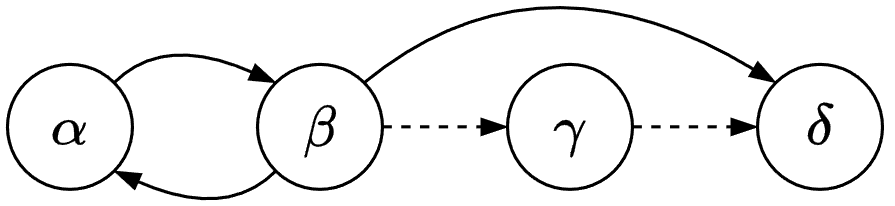}
    \caption{An example QBAF (base scores omitted).}
    \label{fig_1}
\end{figure}

\begin{example}
\label{emp1}
    Consider the QBAF in Figure~\ref{fig_1}. $\gamma$ is disconnected from $\beta$ because there is no path from $\gamma$ to $\beta$; $\gamma$ is single-path connected to $\delta$ via path $p_{\gamma \mapsto \delta}=\langle(\gamma,\delta)\rangle$; $\alpha$ is multi-path connected to $\beta$, because there are infinitely many paths from $\alpha$ to $\beta$
    , 
    namely $p'_{\alpha \mapsto \beta}=\langle(\alpha,\beta)\rangle$, $p''_{\alpha \mapsto \beta}=\langle(\alpha,\beta), (\beta,\alpha), (\alpha,\beta)\rangle$ and so on.
\end{example}

Inspired by~\cite{KR2023_57,AAE_ECAI}, we define  polarity to 
characterize the influence 
between arguments according to their connectivity and the number of attacks on paths between them.
\begin{definition}
\label{def_polarity}
The \emph{polarity} from $\beta$ to $\alpha$ ($\alpha,\!\beta \!\!\in\!\! \mathcal{A}, \alpha \!\!\neq\! \!\beta$) is:
\begin{itemize}
    \item \emph{neutral} ($\beta$ is \emph{neutral} to $\alpha$) iff $\beta$ is disconnected from $\alpha$;
    \item \emph{positive} ($\beta$ is \emph{positive} to $\alpha$) iff
     $P_{\alpha \mapsto \beta} \neq \emptyset$ and
    for every path $p_{\alpha \mapsto \beta} \in P_{\beta \mapsto \alpha}$, $\left| p_{\beta \mapsto \alpha} \cap \mathcal{R}^{-} \right|$ is even;
    \item \emph{negative} ($\beta$ is \emph{negative} to $\alpha$) iff 
    $P_{\beta \mapsto \alpha} \neq \emptyset$ and
    for every path $p_{\beta \mapsto \alpha} \in P_{\beta \mapsto \alpha}$, $\left| p_{\beta \mapsto \alpha} \cap \mathcal{R}^{-} \right|$ is odd;
    
    \item \emph{unknown} ($\beta\!$ is \emph{unknown} to $\alpha$) iff $\exists p'_{\beta \mapsto \alpha}\!\! \in \!\! P_{\beta \mapsto \alpha}$ such that $\!\left| p'_{\beta \mapsto \alpha}\! \cap \! \mathcal{R}^{-} \!\right|\!$ is even and $\exists p''_{\beta \mapsto \alpha} \!\! \in \!\! P_{\beta \mapsto \alpha}$ such that $\!\left| p''_{\beta \mapsto \alpha}\! \cap \! \mathcal{R}^{-} \!\right|\!$ is odd.
\end{itemize}
\end{definition}

So, if $\beta$ 
is single-path connected to $\alpha$, then 
it is 
positive or negative to $\alpha$; if $\beta$ is multi-path connected to $\alpha$, unless 
it is positive or negative to $\alpha$ 
on every path, 
it is unknown to $\alpha$
.

\begin{example}
\label{emp2}
In Figure~\ref{fig_1}, 
$\gamma$ is neutral to $\beta$; 
$\gamma$ is positive to $\delta$ as there is only one path $p_{\gamma \mapsto \delta}$ 
with $0$ (even number) attacks;
$\alpha$ is negative to $\beta$ since, although there are infinitely many paths from $\alpha$ to $\beta$, each path 
has an odd number of attacks;
$\beta$ is unknown to $\delta$ as for one path $p'_{\beta \mapsto \delta}=\langle (\beta,\delta) \rangle$, it has 1 attack, while for another
path $p''_{\beta \mapsto \delta}=\langle (\beta,\gamma),(\gamma,\delta) \rangle$, it has 0 attacks. Hence, $\beta$ is unknown to $\delta$. 
\end{example}
We can restrict the update direction of arguments based on their polarity
assuming that 
$\sigma$ respects
\emph{directionality} and \emph{monotonicity}.
Directionality~\cite{amgoud2016evaluation} states that 
the strength of an argument depends only on its incoming edges.

\begin{definition}
\label{def_directionality}
A semantics $\sigma$ satisfies \emph{directionality} iff, for any $\mathcal{Q}$ and $\mathcal{Q}'\!\!=\!\!\langle \mathcal{A}', \mathcal{R^-}', \mathcal{R^+}', \tau' \rangle$ such that $\mathcal{A}=\mathcal{A}'$, $\mathcal{R^-} \subseteq \mathcal{R^-}'$, and $\mathcal{R^+} \subseteq \mathcal{R^+}'$, the following holds: for any $\alpha,\beta,\gamma \in \mathcal{A}$, let $\sigma_{\mathcal{Q}'}(\gamma)$ denote the strength of $\gamma$ in $\mathcal{Q}'$, if $\mathcal{R^-}' \cup \mathcal{R^+}'=\mathcal{R^-} \cup \mathcal{R^+} \cup \{(\alpha,\beta)\}$ and 
$P_{\beta \mapsto \gamma} = \emptyset$, then $\sigma(\gamma)=\sigma_{\mathcal{Q}'}(\gamma)$.

\end{definition}
\begin{proposition}
\label{prop_direction}
If a semantics $\sigma$ satisfies directionality, then 
for any $\alpha,\beta \!\in \! \mathcal{A}$ such that $\beta$ is neutral to $\alpha$ and for any $\tau'$ such that $\tau'(\beta)$ is an arbitrary value from $[0,1]$ and $\tau'(\gamma) = \tau(\gamma)$ for all 
$\gamma \!\in\! \mathcal{A} \!\setminus \{\beta\}$, $\sigma_{\tau'}(\alpha) 
= \sigma(\alpha)$ always holds.
\end{proposition}
\begin{proposition}
\label{prop_direction_satisfaction}
QE, REB, DF-QuAD 
 satisfy directionality.
\end{proposition}

For semantics satisfying directionality, 
by Proposition~\ref{prop_direction}, we could just maintain the base scores of \textbf{neutral} arguments (wrt. the topic argument) when identifying counterfactuals.

Monotonicity states that increasing (decreasing) the base score of a single-path connected supporter (attacker) will not decrease (increase) the final strength of the topic argument.
\begin{definition}
\label{def_monotonicity}
A semantics $\sigma$ satisfies \emph{monotonicity} iff 
for any $\alpha,\beta \!\in\! \mathcal{A}$ such that $\beta$ is single-path connected to $\alpha$,
for any $\tau_1,\!\tau_2$ 
with $\tau_1(\beta) \!\leq \!\tau_2(\beta)$ and $\tau_1(\gamma) \!\!=\!\! \tau_2(\gamma)$ for all 
$\gamma \!\!\in \!\!\mathcal{A} \!\setminus \!\{\beta\}$:
\begin{itemize}
    \item if $(\beta,\alpha) \in \mathcal{R^{-}}$, then $\sigma_{\tau_1}(\alpha) \geq \sigma_{\tau_2}(\alpha)$;
    \item if $(\beta,\alpha) \in \mathcal{R^{+}}$, then $\sigma_{\tau_1}(\alpha) \leq \sigma_{\tau_2}(\alpha)$.
\end{itemize}
\end{definition}

\begin{proposition}
\label{prop_mono}
If a semantics $\sigma$ satisfies monotonicity, then for any $\tau_1,\tau_2$ such that $\tau_1(\beta) \leq \tau_2(\beta)$ and $\tau_1(\gamma) = \tau_2(\gamma)$ for all 
$\gamma \in \mathcal{A} \setminus \{\beta\}$:
\begin{itemize}
    \item if $\beta$ is negative to $\alpha$, then $\sigma_{\tau_1}(\alpha) \geq \sigma_{\tau_2}(\alpha)$;
    \item if $\beta$ is positive to $\alpha$, then $\sigma_{\tau_1}(\alpha) \leq \sigma_{\tau_2}(\alpha)$.
\end{itemize}
\end{proposition}

\begin{proposition}
\label{prop_monotonicity_satisfaction}
QE, REB, DF-QuAD satisfy monotonicity.
\end{proposition}

For semantics satisfying monotonicity, by Prop.~\ref{prop_mono}, we can increase a topic argument's the strength by increasing (decreasing) the base scores of \textbf{positive}
(\textbf{negative}) arguments
.

\begin{figure}[t]
    \centering
    \includegraphics[width=0.98\columnwidth]{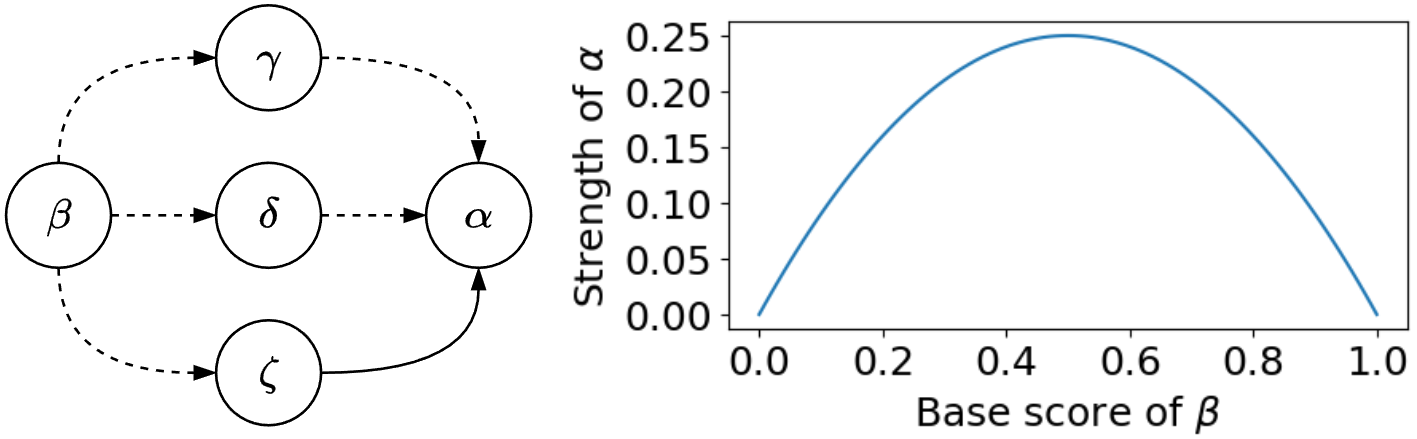}
    \caption{A QBAF evaluated by the DF-QuAD semantics (inspired by \protect\cite{kampik2024contribution}).}
    \label{fig_multi_mono}
\end{figure}


We now consider unknown arguments. Since their base scores may not be globally monotonic to the strength of a topic argument, we cannot simply decide an invariant updating direction for their base scores.
For example, in Figure~\ref{fig_multi_mono} where the base scores of all arguments are 0, with the increase of $\beta$'s base score from 0 to 1 while all others remain the same, the final strength of $\alpha$ displays a non-monotonic effect, increasing initially and then decreasing.
To overcome this challenge, we introduce the \emph{difference quotient} from one argument to another, enabling us to capture the average changing rate of the strength wrt. the base score within 
an 
interval, approximately reflecting local monotonicity at the current base score. 
Then, the difference quotient can act as indicator for the updating direction of \textbf{unknown} arguments.
\begin{definition}
\label{def_diff_quotient}
For any $\alpha,\beta \in \mathcal{A}$, 
given a 
constant $h \in [-1,0) \cup (0,1]$ and 
a base score function $\tau'$ such that $\tau'(\beta)=\tau(\beta)+h \in [0,1]$ and $\tau'(\gamma) = \tau(\gamma)$ for all 
$\gamma \in \mathcal{A} \setminus \{\beta\}$, 
the \emph{difference quotient} from $\beta$ to $\alpha$ is
: $$\Delta^{h}_{\beta \mapsto \alpha}=\frac{\sigma_{\tau'}(\alpha)-\sigma(\alpha)}{h}.$$
\end{definition}
As an illustration, if the difference quotient from an unknown argument to the topic argument is greater (less) than $0$, we increase (decrease) its base score when the desired strength is higher than the current. However, if the difference quotient is exactly equal to $0$, we do not update. 
To highlight the compatibility of the difference quotient with our previous ideas, let us note that 
neutral arguments always have 0 difference quotient, while positive (negative) arguments always have positive (negative) 
quotients.
\begin{proposition}[Sign Invariance]
\label{prop_diff_polarity_relationship}
For any $\alpha,\beta \in \mathcal{A}$, 
for any $h \in [-1,0) \cup (0,1]$,
if a semantics $\sigma$ satisfies directionality and monotonicity, then 
\begin{itemize}
    \item if $\beta$ is neutral to $\alpha$, $\Delta^{h}_{\beta \mapsto \alpha} = 0$;
    \item if $\beta$ is positive to $\alpha$, $\Delta^{h}_{\beta \mapsto \alpha} \geq 0$;
    \item if $\beta$ is negative to $\alpha$, $\Delta^{h}_{\beta \mapsto \alpha} \leq 0$.
\end{itemize}
\end{proposition}

After determining the updating directions for all four types of arguments, we consider updating the magnitude.
To do so, we use a \emph{priority} strategy 
that assigns higher updating magnitude to arguments closer to the topic argument in terms of path length.
We expect that an appropriate priority can help identify cost-effective counterfactuals in terms of $L_p$-norm distance, 
under the hypothesis that updating closer arguments is more efficient in achieving the desired strength.
We empirically verify this 
hypothesis in Section~\ref{sec_evaluation}.



We define \emph{priority} as the reciprocal of the shortest path length from one argument to another, which is at most 1 for the attackers or supporters
.
\begin{definition}
\label{def_priority}
For any $\alpha,\beta \in \mathcal{A}$, $\alpha \neq \beta$: 
\begin{itemize}
    \item if $\beta$ is disconnected from $\alpha$, then the \emph{priority} from $\beta$ to $\alpha$ is 
    $0$;
    \item if $\beta$ is single-path or multi-path connected to $\alpha$ via $n (n \geq 1)$ paths $p^{1}_{\beta \mapsto \alpha},\cdots,p^{n}_{\beta \mapsto \alpha} \in P_{\beta \mapsto \alpha}$, then the \emph{priority} from $\beta$ to $\alpha$ is 
    $1/min\{ |p_{\beta \mapsto \alpha}| \mid p_{\beta \mapsto \alpha} \in P_{\beta \mapsto \alpha} \}$.
\end{itemize}
\end{definition}

\begin{example}
\label{}
Consider the QBAF in Figure~\ref{fig_1}. 
Let us first consider the priority from $\alpha$ to $\delta$. There are infinitely many paths from $\alpha$ to $\delta$. Among  them, path $p_{\alpha \mapsto \delta}\!=\!\langle (\alpha,\beta),(\beta,\delta) \rangle$ has the minimal length 2, thus, the priority from $\alpha$ to $\delta$ is $0.5$. 
Since the minimal length of paths from $\beta$ to $\delta$ and from $\gamma$ to $\delta$ are both $1$, the priorities are both $1$. 
\end{example}

\subsection{Algorithms 
}
\label{subsec:algo}
Algorithm~\ref{algo_polarity} computes the polarity from $\alpha$ to $\beta$ ($\alpha \neq \beta$) in three steps. 
Firstly, we compute all the non-cyclic paths from $\alpha$ to $\beta$ with \emph{Depth-First Search (DFS)} and store them in a set of paths called $P_{\alpha \mapsto \beta}$. If there is no path, then the polarity is neutral.
Secondly, all the nodes in all the paths are checked for 
cycles with the function \emph{find\_cycles} \footnote{See \url{https://github.com/XiangYin2021/CE-QArg} for details. A quicker implementation could involve applying \cite{johnson1975finding}, which is able to find all the elementary cycles of a directed graph in time bounded by O((n+e)(c+1)), where n, e, and c are the number of nodes, edges, and elementary cycles (nodes occur once except for the starting node) in the graph, respectively.}, which outputs a set of elementary cycles.
If any of the node 
is part of a cycle, then we check 
whether 
each cycle contains an odd number of attacks. If this is the case, then the polarity from $\alpha$ to $\beta$ is unknown because the cycle will contain both odd and even numbers of attacks by going through the cycle an odd or even number of times.
For instance, suppose a QBAF consists of two arguments $\alpha$ and $\beta$, where $\alpha$ attacks $\beta$, and $\beta$ supports $\alpha$. In this case, path $\langle (\alpha,\beta) \rangle$ is negative while $\langle (\alpha,\beta), (\beta,\alpha), (\alpha,\beta) \rangle$ is positive, thus $\alpha$ is unknown to $\beta$.
Finally, we check the number of attacks on every path: if all the paths contain an odd (even) number of attacks, $\alpha$ is negative (positive) to $\beta$, and it is unknown otherwise.
\begin{algorithm}[tb]
    \caption{Polarity Computation Algorithm}
    \label{algo_polarity}
    \textbf{Input}: A QBAF $\mathcal{Q}$, two arguments $\alpha,\beta \in \mathcal{A}$ \\
    \textbf{Output}: The polarity from $\alpha$ to $\beta$
    \begin{algorithmic}[1] 
        \STATE $P_{\alpha \mapsto \beta} \leftarrow DFS(\mathcal{Q}, \alpha, \beta)$ 
        \IF{$P_{\alpha \mapsto \beta} == \emptyset$}
        \RETURN $-2$ //neutral
        \ENDIF
        \FOR{$p_{\alpha \mapsto \beta}$ in $P_{\alpha \mapsto \beta}$}
            \FOR{$node$ in $p_{\alpha \mapsto \beta}$}
                \STATE $cycles \leftarrow find\_cycles(\mathcal{Q}, node)$
                \FOR{$cycle$ in $cycles$}
                    \IF{$cycle$ contains odd number attacks}
                        \RETURN $0$ // unknown
                        \ENDIF
                    \ENDFOR
            \ENDFOR
        \ENDFOR
        \STATE $polarity \leftarrow [\ ]$ // empty array
        \FOR{$p_{\alpha \mapsto \beta}$ in $P_{\alpha \mapsto \beta}$}
            \IF{$p_{\alpha \mapsto \beta}$ contains odd number attacks}
                \STATE $polarity.append(-1)$ // negative
            \ELSIF{$p_{\alpha \mapsto \beta}$ contains even number attacks}
                \STATE $polarity.append(1)$ // positive
            \ENDIF
        \ENDFOR
        \IF{all items in $polarity == -1$}
            \RETURN $-1$ // negative
        \ELSIF{all items in $polarity == 1$}
            \RETURN $1$ // positive
        \ELSE{}
            \RETURN $0$ // unknown
        \ENDIF
    \end{algorithmic}
\end{algorithm}

Algorithm~\ref{algo_priority} computes the priority from $\alpha$ to $\beta$. 
We define the 
self-priority of an argument as a constant greater than 1. 
Thus an argument has the highest priority to itself than any others.
Next, we perform a 
DFS to compute all the non-cyclic paths from $\alpha$ to $\beta$ and return the reciprocal of the shortest path length.
\begin{algorithm}[tb]
    \caption{Priority Computation Algorithm}
    \label{algo_priority}
    \textbf{Input}: A QBAF $\mathcal{Q}$, two arguments $\alpha,\beta \in \mathcal{A}$, a constant $c$ \\
    \textbf{Output}: The priority from $\alpha$ to $\beta$
    \begin{algorithmic}[1] 
        \IF{$\alpha == \beta$}
            \RETURN $c$
        \ENDIF
        \STATE $P_{\alpha \mapsto \beta} \leftarrow DFS(\mathcal{Q}, \alpha, \beta)$ 
        \RETURN $1/min\{ length(p_{\alpha \mapsto \beta}) \mid p_{\alpha \mapsto \beta} \in P_{\alpha \mapsto \beta} \}$
    \end{algorithmic}
\end{algorithm}



Algorithm~\ref{algo_tailored} (which we call \emph{\textbf{CE-QArg}} for Counterfactual Explanations for Quantitative bipolar Argumentation frameworks) is an iterative updating algorithm for identifying valid and cost-effective $\delta$-approximate counterfactuals
. CE-QArg essentially involves determining the direction and magnitude of arguments.
For brevity, we assume the current strength of the topic argument is less than the desired one.
Firstly, we compute the polarity and priority for each argument with the function \emph{func\_polarity} and \emph{func\_priority} from Algorithm~\ref{algo_polarity} and \ref{algo_priority}, respectively.
Secondly, the key part of this algorithm is to determine the updating direction in every updating iteration. 
For this, we need an \emph{update} list to record the updating direction in each iteration.
For positive, negative and neutral arguments (lines 5-10), we only need to identify the updating direction once since they remain invariant, whereas, for unknown arguments (lines 13-20), we need to compute their updating direction in every iteration by difference quotient using the function \emph{func\_dquo}, which can be intuitively implemented by Definition~\ref{def_diff_quotient} (so we omit the details).
Once the updating direction of every argument is determined in an iteration, we update the base scores of all arguments all in one go by a small step multiplied by their priority (line 22) and make sure they are within the bounds. We assume the step is small enough that the interactions among arguments can be neglected.
We repeat this procedure iteratively until the current strength reaches the desired strength, after which we return a possible counterfactual (lines 11-24).

\begin{algorithm}[tb]
    \caption{CE-QArg}
    \label{algo_tailored}
    \textbf{Input}: A QBAF $\mathcal{Q}$, a gradual semantics $\sigma$, a topic argument $\alpha^*$ and a desired strength $s^*$ for $\alpha^*$\\
    \textbf{Parameter}: An updating step $\varepsilon$, a change $h$\\
    \textbf{Output}: A counterfactual $\tau^*$
    \begin{algorithmic}[1] 
        \STATE $update, polarity, priority \leftarrow \{\},\{\},\{\}$ //dictionaries
        
        \FOR{$\alpha$ in $\mathcal{A}$}
            \STATE $polarity[\alpha] \leftarrow func\_polarity(\alpha,\alpha^*)$
            \STATE $priority[\alpha] \leftarrow func\_priority(\alpha,\alpha^*)$
            \IF {$polarity[\alpha] ==-2$} 
                \STATE $update[\alpha] \leftarrow 0$  // neutral
            
            \ELSIF {$polarity[\alpha] == 1$}
                \STATE $update[\alpha] \leftarrow 1$ // positive
    
            \ELSIF {$polarity[\alpha] == -1$}
                \STATE $update[\alpha] \leftarrow -1$ // negative
            \ENDIF
        \ENDFOR
        
        \WHILE{$\sigma(\alpha^*) < s^*$}
            \FOR{$\alpha$ in $\mathcal{A}$}
                \IF {$polarity[\alpha] == 0$}
                    \STATE $dquo[\alpha] \leftarrow func\_dquo(\alpha,\alpha*,h)$
                    \IF {$dquo[\alpha] > 0$}
                        \STATE $update[\alpha] \leftarrow 1$
                    \ELSIF {$dquo[\alpha] < 0$}
                        \STATE $update[\alpha] \leftarrow -1$
                    \ELSE 
                        \STATE $update[\alpha] \leftarrow 0$
                    \ENDIF 
                \ENDIF 
            \ENDFOR
            \FOR {$\alpha$ in $\mathcal{A}$}
                \STATE $\tau^*(\alpha) \leftarrow max(0, min(1,\tau^*(\alpha)+update[\alpha] \cdot \varepsilon \cdot priority[\alpha]))$
            \ENDFOR
            \STATE compute $\sigma(\alpha^*)$
        \ENDWHILE
        \STATE \textbf{return} $\tau^*$
    \end{algorithmic}
\end{algorithm}

\section{Formal Properties for Explanations}
\label{sec_properties}
We study the properties for counterfactuals, with a focus on the neutral, negative, and positive arguments. 
Here, we assume the gradual semantics considered satisfy both directionality and monotonicity. 
Note that the properties of explanations are distinct from those of semantics, despite the satisfaction of the former being dependent on the latter.

Existence~\cite{vcyras2022dispute,kampik2024contribution} is a commonly considered property for explanations. It says that if the strength of an argument differs from its base score, then there must exist an argument that caused the change.
\emph{Alteration Existence} states that if a valid counterfactual increases the strength of the topic argument, then there must exist a positive (negative) argument whose base score is also increased (decreased) in the counterfactual whenever the QBAF does not have any unknown arguments.
\begin{proposition}[Alteration Existence]
Given 
a valid counterfactual $\tau^*$ (for the strong/$\delta-$approximate/weak counterfactual problem)
,
a topic argument $\alpha^* \in \mathcal{A}$ such that $\tau^*(\alpha^*)=\tau(\alpha^*)$,
and $\nexists \beta \in \mathcal{A}$ $(\beta \neq \alpha^*)$ such that $\beta$ is unknown to $\alpha^*$:
\begin{enumerate}
    \item If $\sigma(\alpha^*) \!\neq\! \sigma_{\tau^*}(\alpha^*)$, then 
    $\exists \!\gamma \!\in \!\mathcal{A}$ such that $\tau(\gamma) \!\neq\! \tau^*(\gamma)$;
    \item If $\sigma(\alpha^*) < \sigma_{\tau^*}(\alpha^*)$, then either $\exists \gamma \in \mathcal{A}$ such that $\gamma$ is positive to $\alpha^*$ and $\tau(\gamma) \leq \tau^*(\gamma)$ or $\exists \gamma \in \mathcal{A}$ such that $\gamma$ is negative to $\alpha^*$ and $\tau(\gamma) \geq \tau^*(\gamma)$;
    \item If $\sigma(\alpha^*) > \sigma_{\tau^*}(\alpha^*)$, then either $\exists \gamma \in \mathcal{A}$ such that $\gamma$ is positive to $\alpha^*$ and $\tau(\gamma) \geq \tau^*(\gamma)$ or $\exists \gamma \in \mathcal{A}$ such that $\gamma$ is negative to $\alpha^*$ and $\tau(\gamma) \leq \tau^*(\gamma)$.
\end{enumerate}
\end{proposition}


Given that validity is fundamental 
for counterfactuals, we  introduce two properties 
associated with validity.

For 
attribution-based explanations for QBAFs, it is interesting to explore the effects of removing or nullifying an argument (by setting its base score to $0$ -- e.g. see removal-based contribution functions in \cite{kampik2024contribution} and
\emph{agreement} in \cite{AAE_ECAI}).
For counterfactual explanations, it is essential to consider the validity of a counterfactual if it is perturbed
.
Combining both ideas, we propose a novel property called \emph{nullified-validity}, which examines whether a valid counterfactual remains valid even after nullifying an argument.
For example, nullifying a positive argument in a valid counterfactual could still result in a valid counterfactual if a smaller strength is expected for the topic argument in the weak counterfactual problem.
\begin{proposition}[Nullified-Validity]
Given 
a valid counterfactual $\tau^*$ (for the strong/$\delta-$approximate/weak counterfactual problem)
,
a topic argument $\alpha^* \in \mathcal{A}$ and a desired strength $s^*$ for $\alpha^*$, 
and another base score function $\tau^0$ such that 
for some
$\beta \in \mathcal{A}$ $(\beta \neq \alpha^*)$, $\tau^0(\beta)=0$ and $\tau^0(\gamma) = \tau^*(\gamma)$ for all $\gamma \in \mathcal{A} \setminus \{\beta\}$:
\begin{enumerate}
    \item If $\beta$ is neutral to $\alpha^*$,
    then $\tau^0$ is still 
    valid for the strong/$\delta-$approximate/weak counterfactual problem;
    \item If $\beta$ is negative to $\alpha^*$ and $\sigma_{\tau^*}(\alpha^*) \!\!\geq \!\! s^*$,
    then $\tau^0$ is still 
valid 
for the weak counterfactual problem;
    \item If $\beta$ is positive to $\alpha^*$ and $\sigma_{\tau^*}(\alpha^*) \leq s^*$,
    then $\tau^0$ is still 
    valid for the weak counterfactual problem.
\end{enumerate}
\end{proposition}

In attribution-based explanations for QBAFs, it is also interesting to compare two related explanations and study their properties (e.g., see \emph{monotonicity} in \cite{AAE_ECAI} and \emph{dominance} in \cite{amgoud2017measuring,YIN_RAE_IJCAI}).
We extend this idea to counterfactuals by focusing on the validity property.
\emph{Related-validity} identifies a valid counterfactual by comparing it with another already valid counterfactual without recomputing the strengths of arguments.
To illustrate, suppose we have a valid counterfactual for the weak counterfactual problem, where the strength of the topic argument is as small as desirable. Then we can compare it with another counterfactual: 
if the latter counterfactual has a smaller base score in a positive argument while keeping all other base scores the same, it is still considered valid.

\begin{proposition}[Related-Validity]
Given 
a valid counterfactual explanation $\tau^*$ (for the weak counterfactual problem)
,
a topic argument $\alpha^* \in \mathcal{A}$ and a desired strength $s^*$ for $\alpha^*$ such that $\sigma_{\tau^*}(\alpha^*) \geq s^*$, for every other
base score function $\tau'$ in $\mathcal{Q}_{\tau'}$,
and all $\beta \in \mathcal{A}$ $(\beta \neq \alpha^*)$:
\begin{enumerate}
    \item If $\beta \in \mathcal{A}$ is neutral to $\alpha^*$, $\tau'(\beta) \geq \tau^*(\beta)$ and $\tau'(\gamma) = \tau^*(\gamma)$ for all 
    $\gamma \in \mathcal{A} \setminus \{\beta\}$, then $\tau'$ is also valid (for the weak counterfactual problem);
    \item If $\beta \in \mathcal{A}$ is negative to $\alpha^*$, $\tau'(\beta) \leq \tau^*(\beta)$ and $\tau'(\gamma) = \tau^*(\gamma)$ for all 
    $\gamma \in \mathcal{A} \setminus \{\beta\}$, then $\tau'$ is also a valid counterfactual (for the weak counterfactual problem);
    \item If $\beta \in \mathcal{A}$ is positive to $\alpha^*$, $\tau'(\beta) \geq \tau^*(\beta)$ and $\tau'(\gamma) = \tau^*(\gamma)$ for all 
    $\gamma \in \mathcal{A} \setminus \{\beta\}$, then $\tau'$ is also a valid counterfactual (for the weak counterfactual problem).
\end{enumerate}
\end{proposition}

\section{Evaluations}
\label{sec_evaluation}
We show  effectiveness (Experiment 1), scalability (Experiment 2) and robustness (Experiment 3) of CE-QArg.
In this section, we focus on the $\delta-$approximate counterfactual problem and valid explanations therefor with $\delta=0.1$.

\paragraph{Settings}
We conducted experiments separately using acyclic and cyclic QBAFs.
For \textbf{acyclic} QBAFs, we generated tree-like QBAFs as they occur in many applications of QBAFs (e.g. see \cite{kotonya2019gradual,cocarascu2019extracting,chi2021optimized}). We created full binary, ternary, and quaternary trees with different widths (2, 3, and 4) and depths (from 1 to 8) where each edge was randomly set to an attack or a support.
The topic argument was set as the root of the tree.
To improve the credibility of the results and reduce the impact of random errors, we randomly created each tree-like QBAF with a specified width and depth 100 times with different base scores in $[0,1]$ over different arguments.
For \textbf{cyclic} QBAFs, we created varying numbers of arguments (from 100, 200, to 1000), each repeated 100 times with random attacks or supports, random base scores in $[0,1]$, and a randomly designated topic argument. 
The argument-relation ratio was set as 1:1 to avoid dense QBAFs which can impact the explainability because of their high structural complexity, making them difficult to comprehend and thus less suitable 
for explainability. 

We report on experiments with the QE semantics\footnote{
We give results with DF-QuAD and REB in \url{https://arxiv.org/abs/2407.08497}.}. 
We set the updating step $\varepsilon$ to 0.01 in each iteration. Besides, we used both $L_1$ and $L_2$-norm distance as the metric for cost.

\subsection{Experiment 1: Effectiveness}
We show the effectiveness of CE-QArg by conducting ablation studies on polarity and priority. 
We first propose the Baseline method (BL) based on Proposition~\ref{prop_diff_polarity_relationship}, which directly computes the difference quotient for all arguments as the updating indicators without considering their polarity and priority.
We then separately applied priority or polarity on the BL to show their individual efficacy (denoted as BL+pri and BL+pol). 
Finally, we showed the performance of our CE-QArg which incorporates both polarity and priority. 
We evaluated CE-QArg on validity, $L_1$, $L_2$-norm distance, and runtime. 

\begin{table}[h]
\caption{Ablation studies for polarity and priority on \textbf{acyclic/cyclic} QBAFs: Comparison of average validity, $L_1$, $L_2$-norm distance, and runtime over 100 random generated acyclic full binary tree-like QBAFs with a depth of 7 and cyclic QBAFs with 100 arguments and 100 relations.\protect\footnotemark (The best results are shown in bold.)}
\label{table_Expr_ablation_study_acyclic}
    \centering
    \small
    \begin{tabular}{lcccc}
    \hline
        ~ & Validity  & $L_1$ & $L_2$ & Runtime (s)  \\ \hline
        BL & \textbf{1.00}/0.78 & 16.52/17.75 & 1.65/1.87 & 1.52/79.57  \\ 
        BL+pri & \textbf{1.00}/0.93 & \textbf{4.04}/\textbf{0.51} & \textbf{0.49}/\textbf{0.30} & 1.16/32.30  \\ 
        BL+pol & \textbf{1.00}/\textbf{1.00} & 22.03/1.42 & 2.05/0.46 & 0.02 /1.58 \\ 
        Ours & \textbf{1.00}/\textbf{1.00} & 4.33/0.54 & 0.50/\textbf{0.30} & \textbf{0.01}/\textbf{0.80}  \\ \hline
    \end{tabular}
\end{table}
\footnotetext{We show the effectiveness on larger QBAFs in \url{https://arxiv.org/abs/2407.08497}.}


We first discuss the results of acyclic QBAFs (left side of slash in Table~\ref{table_Expr_ablation_study_acyclic}).
All methods had the best validity of $1.00$.
The use of priority is expected to shorten the $L_1$ and $L_2$-norm distance by updating the arguments close to the topic argument with a larger step. 
The results of BL and BL+pri showed that applying priority resulted in a 75.5\% and 70.3\% decrease in $L_1$ and $L_2$-norm distance, respectively, which also reduced the runtime by 23.7\% in the meanwhile. These findings are in line with our 
hypothesis 
in that priority enables reaching a desired strength more cost-effectively, making the distance shorter and thus requiring fewer iterations.
Utilizing polarity is expected to lower the runtime by computing the polarity only once for neutral, positive, and negative arguments in a QBAF.
The results of BL and BL+pol showed that applying polarity significantly decreased the runtime by 98.7\% as expected because the BL computes the difference quotient in every iteration for every argument, which causes the runtime wastage. 
For our CE-QArg, we see that the runtime is better than that of any of the previous three algorithms
. 
Compared to BL+pri, although the $L_1$ and $L_2$-norm distance is slightly longer in our CE-QArg, the runtime is substantially decreased.

The results of cyclic QBAFs (right side of the slash) 
are similar.
We can observe that using priority alone decreased the $L_1$ and $L_2$-norm distance by 97.1\% and 84.0\%, respectively; and solely applying polarity significantly decreased the runtime by 98.0\%. Finally, applying both priority and polarity can yield the desired counterfactual explanations in terms of the $L_1$ and $L_2$-norm distance and shorten the algorithm runtime.
However, it is interesting to note that the validity was violated without applying polarity (BL and BL+pri).
This is probably because of the computation error of arguments' strength, especially for cyclic QBAFs where an analytical value does not generally exist. Then, the strength error will cause the wrong difference quotient thus the wrong updating direction. As a result, the current strength just oscillates around the desired strength. The invalidity of counterfactuals also explains why the $L_1$ and $L_2$-norm distance performs better than others in BL+pri.

Overall, the ablation studies show the effectiveness of priority and polarity on both acyclic and cyclic QBAFs, which allows finding valid and cost-effective counterfactuals.



\subsection{Experiment 2: Scalability}
\begin{figure}[h]
    \centering
    \includegraphics[width=1.0\columnwidth]{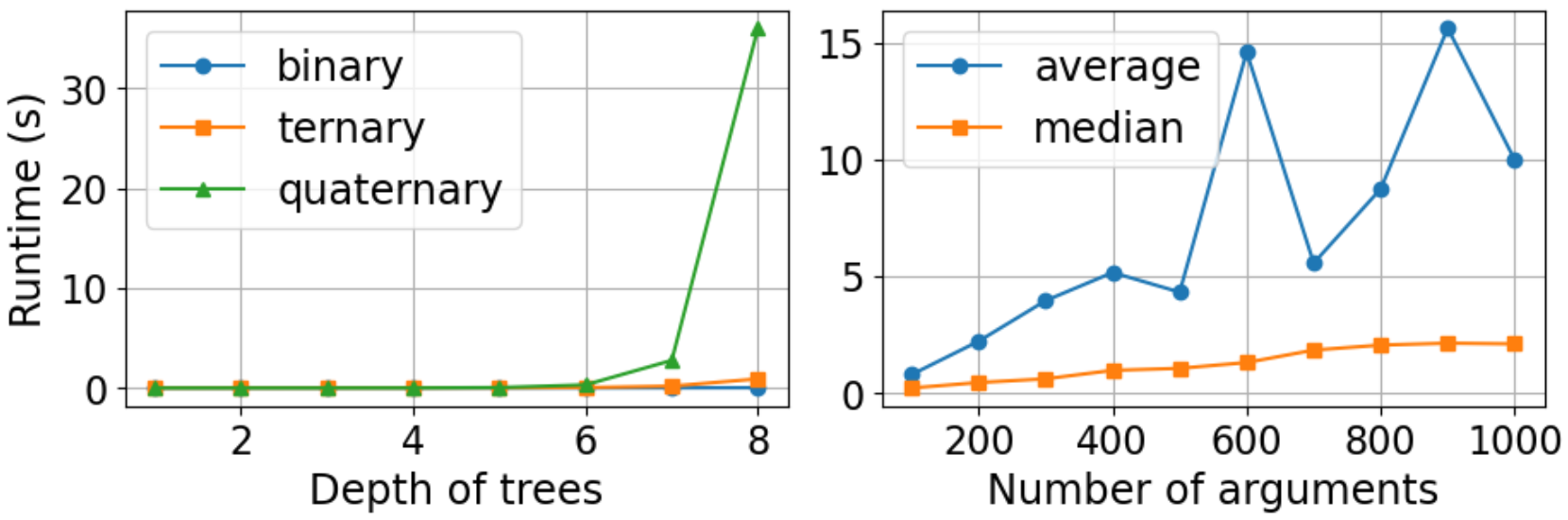}
    \caption{Scalability evaluation for CE-QArg on acyclic (left) and cyclic (right) QBAFs: comparison of average runtime over 100 randomly generated acyclic and cyclic QBAFs.}
    \label{fig_scalability_acyclic}
\end{figure}

We evaluated the scalability of CE-QArg on QBAFs of varying sizes. We show both validity and the runtime performance.
First, all tested acyclic and cyclic QBAFs achieved a validity score of $1.00$.
We next present the average runtime for acyclic and cyclic QBAFs with different sizes in Figure~\ref{fig_scalability_acyclic}.


Figure~\ref{fig_scalability_acyclic} (left) shows the runtime for binary, ternary, and quaternary tree-like QBAFs with depths ranging from 1 to 8.
Notably, binary and ternary trees have a runtime less than 1s across all depths.
However, the runtime for quaternary trees increases sharply from depth 7 (2.76s) to depth 8 (36.00s), which is expected since the number of arguments increases substantially from $(4^7-1)/3=5,461$ to $(4^8-1)/3=21,845$.
In Figure~\ref{fig_scalability_acyclic} (right), we observe the average runtime (blue line) increases as the number of arguments in QBAFs increases.
However, there are two outliers at the number of 600 and 900 which may be due to the randomness of the QBAF generation. Therefore, we added the median runtime
, which shows a stable rising trend as expected.
Note that the runtime line plot can vary dramatically with different densities of QBAFs because CE-QArg involves the traverse of the QBAF using DFS when computing the polarity and priority. This could be improved by adding pruning strategies while traversing to reduce the runtime cost: we leave this improvement to future work.
In summary, both acyclic and cyclic QBAFs exhibit reasonable runtime performance, demonstrating good scalability of CE-QArg.

\subsection{Experiment 3: Robustness}
\begin{figure}[h]
    \centering
    \includegraphics[width=0.6\columnwidth]{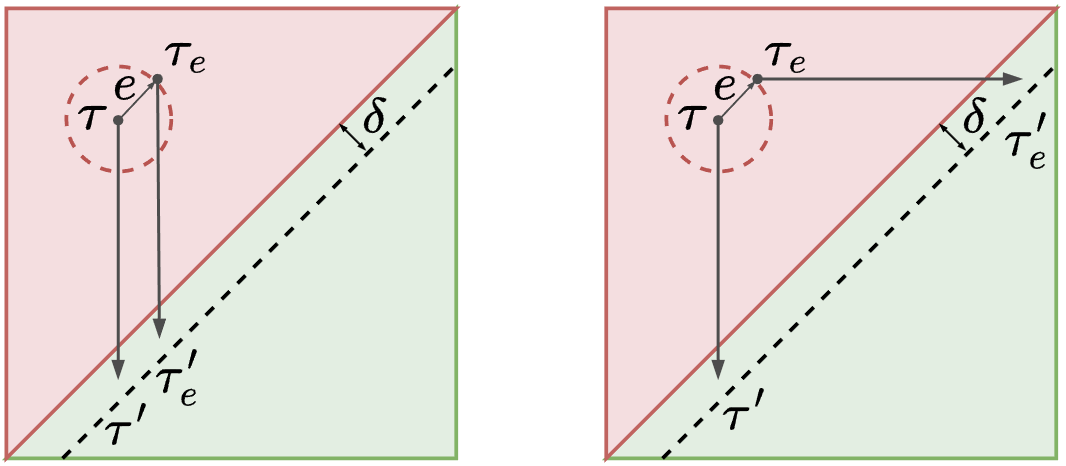}
    \caption{Comparison of robust (left) and non-robust (right) counterfactual explanation methods against input perturbations.}
    \label{fig_robust}
\end{figure}

Robustness is a crucial and commonly considered metric for counterfactual explanation methods~\cite{artelt2021evaluating,jiang_provably,jiang_aaai,jiang_survey}.
An explanation method is robust against input perturbations if similar inputs leading to the same outputs give rise to similar explanations. 
For instance, if two loan applicants with similar conditions are rejected, they should obtain similar counterfactuals.
However, non-robust methods may generate completely different counterfactuals for similar rejected applicants (see an example in Figure 1(b) of~\cite{slack2021counterfactual}), which is unfair as they have different updating costs to obtain the desired decision.
As an illustration, the counterfactual explanation method in Figure~\ref{fig_robust} (left) is more robust against input perturbations than that of Figure~\ref{fig_robust} (right) as the new counterfactual is still close to the previous counterfactual after the input is perturbed.
We propose a \emph{robustness against input perturbations} metric which uses the $L_p$-norm distance to evaluate the robustness of CE-QArg under the perturbation of the input base score function.


\begin{metric}
\label{metric_robustness_explanation}
Given 
a perturbation score $e>0$,
two base score functions $\tau$ and $\tau_{e}= \tau(\alpha) + e$ for all $\alpha \in \mathcal{A}$,
and two counterfactual explanation $\tau'$ for $\tau$ and $\tau'_{e}$ for $\tau_{e}$, 
the \emph{robustness against input perturbations} is measured by $d_p(\tau',\tau'_{e})$.
\end{metric}

In addition, inspired by \cite{leofante2023towards,pawelczyk2022probabilistically}, we propose a \emph{robustness against noisy execution} metric, which requires that the generated counterfactual could still lead to similar output (final strength for the topic argument) even if the 
counterfactual is perturbed.
Still taking the loan application as an example, 
robustness against noisy execution ensures that when a rejected applicant is very close to the provided counterfactual, then the output of this counterfactual should also be close to the desired final strength.
We evaluate robustness against noisy execution by the absolute difference between the strength of a topic argument 
obtained with two similar counterfactuals.
\begin{metric}
\label{metric_robustness_output}
Given
a topic argument $\alpha^* \in \mathcal{A}$,
a perturbation score $e$,
and two counterfactual explanations $\tau'$ and $\tau'_{e}= \tau'(\alpha) + e$ for all $\alpha \in \mathcal{A}$,
the \emph{robustness against noisy execution} is measured by $|\sigma_{\tau'_{e}}(\alpha^*) - \sigma_{\tau'}(\alpha^*)|$.
\end{metric}

\begin{figure}[h]
    \centering
    \includegraphics[width=1.0\columnwidth]{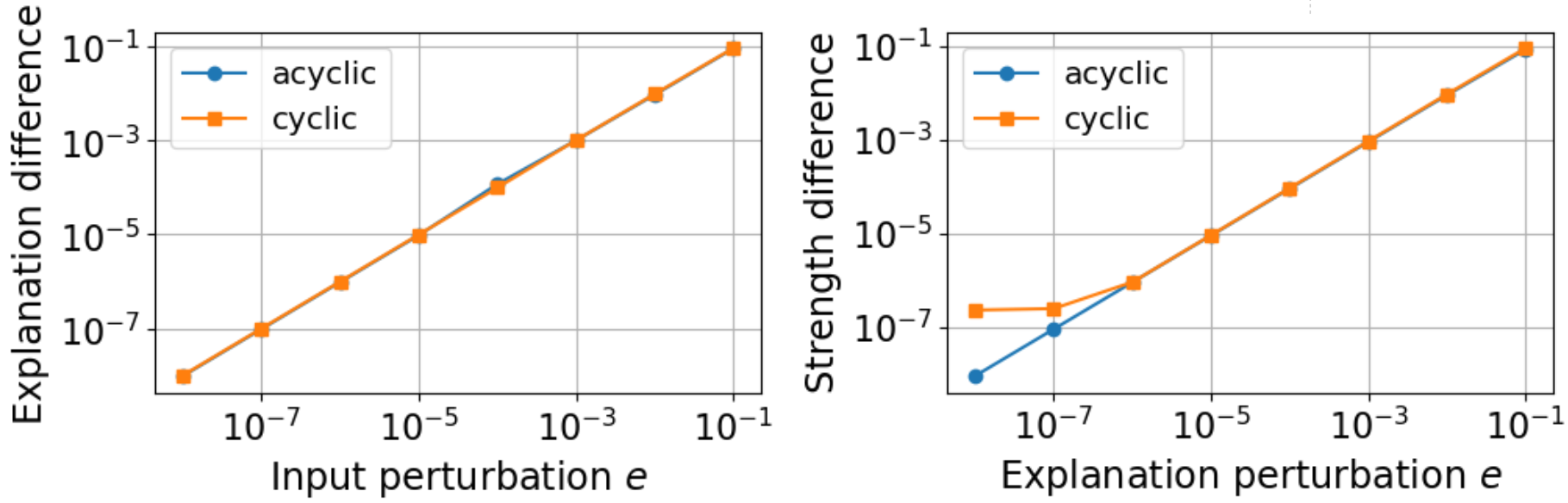}
    \caption{Robustness evaluation of perturbing base score functions (left) and perturbing generated counterfactuals (right) over 100 random generated QBAFs with increasing perturbation $e$ from $10^{-8}$ to $10^{-1}$. Acyclic QBAFs are binary trees with a depth of 7 while cyclic QBAFs contain 100 arguments and relations.\protect\footnotemark}
    \label{fig_robustness}
\end{figure}

We applied Metrics~\ref{metric_robustness_explanation} and \ref{metric_robustness_output} with an increasing perturbation $e$ from $10^{-8}$ to $10^{-1}$ over acyclic and cyclic QBAFs.
Figure~\ref{fig_robustness} (left) shows the robustness against input perturbations through the average explanation difference measured by Metric~\ref{metric_robustness_explanation},
while Figure~\ref{fig_robustness} (right) shows the robustness against noisy execution through the average strength difference of the topic argument measured by Metric~\ref{metric_robustness_output}.
With the increase of $e$,
the explanation difference and strength difference both showed an approximate linear and stable increasing trend, albeit in Figure~\ref{fig_robustness} (right), we observed a bit of unsteadiness at the beginning when $e=10^{-8}$ and $e=10^{-7}$. 
Overall, CE-QArg exhibited robustness against input perturbations and noisy execution on both cyclic and acyclic QBAFs.

\footnotetext{We show the robustness on larger QBAFs in \url{https://arxiv.org/abs/2407.08497}.}

\section{Conclusions}
\label{sec_conclusion}
We formally defined three counterfactual problem variants and discussed their relationships. 
We proposed an iterative algorithm CE-QArg to identify valid and cost-effective counterfactuals.
We discussed some formal properties of our counterfactual explanations and empirically evaluate CE-QArg on random generated QBAFs. Experimental results show that CE-QArg has a desirable performance on effectiveness, scalability, and robustness. 
While the identification of valid and cost-effectiveness counterfactuals still lacks sufficient theoretical guarantees because there is no closed expression for a topic argument's final strength in general (cyclic) graphs, we improved the searching process by applying polarity and priority so that the $L_1$ and $L_2$-norm distance and the runtime decreased significantly compared to the baseline (BL) method. 

There are a few avenues for future work. 
First, it would be interesting to explore identifying counterfactuals in QBAFs when their structure can be changed by adding or removing arguments or edges.
Second, it would be worth exploring identifying counterfactuals for multiple topic arguments simultaneously. However, this would be challenging as it involves the interactive effect among topic arguments.
Third, it would also be interesting to explore the relationship between argument attribution explanations and counterfactual explanations~\cite{kommiya2021towards}. 
Finally, it would be important to carry out case studies and user experiments as explanations should finally help humans understand and make better decisions.

\section*{Acknowledgments}
This research was partially funded by the  European Research Council (ERC) under the
European Union’s Horizon 2020 research and innovation programme (grant
agreement No. 101020934, ADIX) and by J.P. Morgan and by the Royal
Academy of Engineering under the Research Chairs and Senior Research
Fellowships scheme.  Any views or opinions expressed herein are solely those of the authors.

\bibliographystyle{kr}
\bibliography{kr24}

\newpage
\setcounter{page}{1}
\onecolumn
\appendix

\section*{Supplementary Material for\\``CE-QArg: Counterfactual Explanations for\\Quantitative Bipolar Argumentation Frameworks''}
\medskip

\section{Gradual Semantics}
In this section, we will show the computation for the Quadratic Energy (QE), Restricted Euler-based (REB), and Discontinuity-Free Quantitative Argumentation Debates (DF-QuAD) semantics.

The \textbf{QE} semantics is computed as follows:

Aggregation function:
\begin{equation}
\label{equ_qe_agg}
    E_\alpha=\sum_{\left \{ \beta \in \mathcal{A} \mid (\beta,\alpha) \in \mathcal{R^{+}} \right \} }\sigma^{QE}(\beta) - \sum_{\left \{ \beta \in \mathcal{A} \mid (\beta,\alpha) \in \mathcal{R^{-}} \right \} }\sigma^{QE}(\beta)
\end{equation}

Influence function:
\begin{equation}
\label{equ_qe_inf}
    \sigma^{QE}(\alpha)= 
    \begin{cases}
        \tau(\alpha)-\tau(\alpha)\cdot\frac{E_\alpha^2}{1+E_\alpha^2} & if\ E_\alpha \leq 0\\
        \tau(\alpha)+(1-\tau(\alpha))\cdot\frac{E_\alpha^2}{1+E_\alpha^2} & if\ E_\alpha > 0\\
    \end{cases}
\end{equation}

The \textbf{REB} semantics is computed as follows:

Aggregation function:
\begin{equation}
\label{equ_reb_agg}
    E_\alpha=\sum_{\left \{ \beta \in \mathcal{A} \mid (\beta,\alpha) \in \mathcal{R^{+}} \right \} }\sigma^{REB}(\beta) - \sum_{\left \{ \beta \in \mathcal{A} \mid (\beta,\alpha) \in \mathcal{R^{-}} \right \} }\sigma^{REB}(\beta)
\end{equation}

Influence function:
\begin{equation}
\label{equ_reb_inf}
    \sigma^{REB}(\alpha)=1-\frac{1-\tau^2(\alpha)}{1+\tau(\alpha) \cdot e^{E_\alpha}}
\end{equation}

\section{Additional Details for Section~\ref{sec_intro}}

Computation of the importance scores:
According to \cite{kampik2024contribution}, the Shapley-based importance score from $\beta$ to $\alpha$ under the semantics $\sigma$ is defined as follows:
\begin{equation}
\label{equ_shapley}
\phi_{\sigma}^{\alpha}(\beta)\!=\!\!\!\!\!\sum_{\mathcal{S} \subseteq \mathcal{A} \setminus \{\alpha,\beta\}}\!\! \frac{(\left| \mathcal{A} \setminus \{\alpha\}  \right| - \left| \mathcal{S} \right| -1)!\left| \mathcal{S} \right|!}{\left| \mathcal{A} \setminus \{\alpha\} \right|!} \left[ \sigma_{\mathcal{S} \cup \{\beta\}}(\alpha)-\sigma_{\mathcal{S}}(\alpha)\right],
\end{equation}
where $\sigma_{\mathcal{S}}(\alpha)$ denotes the strength of $\alpha$ only when arguments in $\mathcal{S}$ exist in the QBAF, while for arguments not in $\mathcal{S}$, they are deleted from the QBAF when computing the strength of $\alpha$.
In Figure~\ref{fig_loan}, according to Equation~\ref{equ_shapley}, we have
$\phi_{\sigma}^{\alpha}(\beta) = 0.0975$,
$\phi_{\sigma}^{\alpha}(\gamma) = -0.34$,
$\phi_{\sigma}^{\alpha}(\rho) = -0.0525$,
$\phi_{\sigma}^{\alpha}(\zeta) = -0.04$.

\section{Proofs}

\setcounter{proposition}{0}

\begin{proposition}[Solution Existence]
\label{prop_solution_exixtence_sm}
If 
$\sigma$ satisfies s-stability
and $\mathcal{Q}$ is acyclic,
then the trivial counterfactual is a solution to the strong counterfactual problem
.
\end{proposition}
\begin{proof}
Suppose $\alpha^* \in \mathcal{A}$ is the topic argument.
According to Definition~\ref{def_trivial_solution}, we know that the trivial counterfactual explanation is the base score function $\tau'\neq \tau$
such that $\tau'(\alpha^*)=s^*$ and $\tau'(\alpha)=0$ for all
other $\alpha \in \mathcal{A} \setminus \{\alpha^*\}$ in $\mathcal{Q}_{\tau'}$.
Therefore, we have $\sigma(\alpha)=0$ for all the arguments $\alpha$ before $\alpha^*$ in the topological order of the $\mathcal{Q}$ according to the computation of the semantics. Therefore, for all the attackers and supporters of $\alpha^*$, $\sigma(\alpha)=0$. Thus, $\sigma(\alpha^*)=\tau(\alpha^*)=s^*$.
Thus, the trivial counterfactual is a solution to the strong counterfactual problem if $\mathcal{Q}$ is acyclic.
\end{proof}

\begin{proposition}
\label{prop_stability_satisfaction_sm}
QE, REB, and DF-QuAD satisfy s-stability.
\end{proposition}
\begin{proof}

A semantics satisfies s-stability if it satisfies both stability and neutrality.
Since the QE semantics satisfies both stability and neutrality (see \cite{Potyka18} for the proof), it satisfies s-stability.
Analogously, see \cite{amgoud2018evaluation} for the proof of the REB and DF-QuAD semantics.




\end{proof}



\begin{proposition}[Problem Relationships]
\label{prop_rels_sm}
\hfill\par
\begin{enumerate}
\item If a counterfactual is valid for the strong counterfactual problem, then it is also valid for the $\delta-$approximate and weak counterfactual problems.
\item If a counterfactual is valid for the $\delta-$approximate counterfactual problem, then it is also valid for the weak counterfactual problem.
\end{enumerate}
\end{proposition}
\begin{proof}
According to Definition~\ref{def_validity}, if a counterfactual is a solution to the counterfactual problem, then it is valid. Therefore, according to Definition~\ref{def_strong_cfx}, \ref{def_delta_cfx}, and \ref{def_weak_cfx}, we can see the relationship between these problems.
\end{proof}

\setcounter{proposition}{4}

\begin{proposition}
\label{prop_direction_sm}
If a semantics $\sigma$ satisfies directionality, then 
for any $\alpha,\beta \!\in \! \mathcal{A}$ such that $\beta$ is neutral to $\alpha$ and for any $\tau'$ such that $\tau'(\beta)$ is an arbitrary value from $[0,1]$ and $\tau'(\gamma) = \tau(\gamma)$ for all 
$\gamma \!\in\! \mathcal{A} \!\setminus \{\beta\}$, $\sigma_{\tau'}(\alpha) 
= \sigma(\alpha)$ always holds.
\end{proposition}
\begin{proof}
$\beta$ is neutral to $\alpha$ means that $\beta$ is disconnected from $\alpha$ by Definition~\ref{def_polarity}, thus there is no path from $\beta$ to $\alpha$. Therefore, $\beta$ has no influence on $\sigma(\alpha)$ no matter what $\tau(\beta)$ is by Definition~\ref{def_directionality}.
Therefore, $\sigma_{\tau'}(\alpha) \equiv \sigma(\alpha)$ always holds as long as $\tau'(\gamma) = \tau(\gamma)$ for all other $\gamma \in \mathcal{A} \setminus \{\beta\}$.
\end{proof}

\begin{proposition}
\label{prop_direction_satisfaction_sm}
QE, REB, DF-QuAD satisfy directionality.
\end{proposition}
\begin{proof}
See \cite{Potyka18} for the proof of the QE semantics.
See \cite{amgoud2018evaluation} for the proof of the REB and DF-QuAD semantics.
\end{proof}

\begin{proposition}
\label{prop_mono_sm}
If a semantics $\sigma$ satisfies monotonicity, then for any $\tau_1,\tau_2$ such that $\tau_1(\beta) \leq \tau_2(\beta)$ and $\tau_1(\gamma) = \tau_2(\gamma)$ for all 
$\gamma \in \mathcal{A} \setminus \{\beta\}$:
\begin{itemize}
    \item if $\beta$ is negative to $\alpha$, then $\sigma_{\tau_1}(\alpha) \geq \sigma_{\tau_2}(\alpha)$;
    \item if $\beta$ is positive to $\alpha$, then $\sigma_{\tau_1}(\alpha) \leq \sigma_{\tau_2}(\alpha)$.
\end{itemize}
\end{proposition}
\begin{proof}

Let $a_1,a_2,a_3,\ldots,a_n \in \mathcal{A}$, $\beta=a_1$ and $\alpha=a_n$, $(n \geq 2)$.

\textbf{Case 1}: When $a_1$ is positive to $a_n$, there are two sub-cases:

\textbf{Case 1a}: $a_1$ is single-path connected to $a_n$ and there are even number of attacks on the path according to Definition~\ref{def_polarity}.

Suppose $a_1$ is single-path connected to $a_n$ through path $p_{a_1 \mapsto a_n}=\langle(a_1,a_2), (a_2,a_3),\ldots, (a_{n-1},a_n)\rangle $.
Since $a_1$ is single-path connected to $a_n$, for any $a_i (1<i<n)$, $a_i$ is also single-path connected to $a_n$, otherwise $a_1$ is no longer single-path connected to $a_n$. 
Therefore, for any $a_i (1 \leq i<n)$, $a_i$ is single-path connected to $a_n$.

If $\tau_1(a_1) \leq \tau_2(a_1)$, then we have $\sigma_{\tau_1}(a_{1}) \leq \sigma_{\tau_2}(a_{1})$. Then for any $a_i (1 \leq i < n)$, either $\sigma_{\tau_1}(a_i) \leq \sigma_{\tau_2}(a_i)$ or $\sigma_{\tau_1}(a_i) \geq \sigma_{\tau_2}(a_i)$, which will further affect $a_{i+1}$'s strength based on the relation from $a_i$ to $a_{i+1}$. 

Specifically, there are 4 cases that how $a_i$ affects $a_{i+1}$ according to the satisfaction of monotonicity.

\begin{enumerate}
\item $\sigma_{\tau_1}(a_i) \leq \sigma_{\tau_2}(a_i) \wedge (a_i,a_{i+1}) \in \mathcal{R^-} \Rightarrow \sigma_{\tau_1}(a_{i+1}) \geq \sigma_{\tau_2}(a_{i+1})$
\item $\sigma_{\tau_1}(a_i) \leq \sigma_{\tau_2}(a_i) \wedge (a_i,a_{i+1}) \in \mathcal{R^+} \Rightarrow \sigma_{\tau_1}(a_{i+1}) \leq \sigma_{\tau_2}(a_{i+1})$
\item $\sigma_{\tau_1}(a_i) \geq \sigma_{\tau_2}(a_i) \wedge (a_i,a_{i+1}) \in \mathcal{R^-} \Rightarrow \sigma_{\tau_1}(a_{i+1}) \leq \sigma_{\tau_2}(a_{i+1})$
\item $\sigma_{\tau_1}(a_i) \geq \sigma_{\tau_2}(a_i) \wedge (a_i,a_{i+1}) \in \mathcal{R^+} \Rightarrow \sigma_{\tau_1}(a_{i+1}) \geq \sigma_{\tau_2}(a_{i+1})$
\end{enumerate}

Based on these 4 cases, we find that (1) every time when $a_i$ passes an attack, the increase of $a_i$'s strength will cause the decrease of $a_{i+1}$'s strength; (2) every time when $a_i$ passes a support, the increase of $a_i$'s strength will cause the increase of $a_{i+1}$'s strength.

Initially, since $\tau_1(a_1) \leq \tau_2(a_1)$, we have $\sigma_{\tau_1}(a_{1}) \leq \sigma_{\tau_2}(a_{1})$. Then, along this path, every time $a_i$ passes an attack, the strength magnitude relationship for the $a_{i+1}$ differs from that of $a_i$.

Thus, if $a_1$ passes even number of attacks, then the strength magnitude relationship for the $a_{n}$ is the same as that of $a_1$, that is, $\sigma_{\tau_1}(a_{i+1}) \leq \sigma_{\tau_2}(a_{i+1})$.
Therefore, if $a_1$ is positive to $a_n$, then $\sigma_{\tau_1}(a_n) \leq \sigma_{\tau_2}(a_n)$.

\textbf{Case 1b}: $a_1$ is multi-path connected to $a_n$ and each path contains even number of attacks.
In this case, similar to the single-path cases, that $a_1$ will either decrease the strength of attackers of $a_n$ or increase the strength of supporters of $a_n$; or both way.
Therefore, if $a_1$ is positive to $a_n$, then $\sigma_{\tau_1}(a_n) \leq \sigma_{\tau_2}(a_n)$.

\textbf{Case 2}: $\beta$ is negative to $\alpha$. 
Analogously to the proof in case 1.




\end{proof}

\begin{proposition}
\label{prop_monotonicity_satisfaction_sm}
QE, REB, DF-QuAD satisfy monotonicity.
\end{proposition}
\begin{proof}

The proof for the QE semantics:

1. For any $\alpha,\beta \in \mathcal{A}$ such that $\beta$ is a single-path connected attacker to $\alpha$. If $\tau_1(\beta) \leq \tau_2(\beta)$ and $\tau_1(\gamma) = \tau_2(\gamma)$ for all other $\gamma \in \mathcal{A} \setminus \{\beta\}$, then $E_\beta$ still remains the same as the strength of $\beta$'s attackers and supporters does not change. Then, $\sigma_{\tau_1}^{QE}(\beta) \leq \sigma_{\tau_2}^{QE}(\beta)$. 
For $\alpha$, since $\beta$ is the only attacker that the final strength changes, thus, $E_\alpha$ decreases or maintains. Therefore, $\sigma_{\tau_1}(\alpha) \geq \sigma_{\tau_2}(\alpha)$ by Equation~\ref{equ_qe_inf}.

2. For any $\alpha,\beta \in \mathcal{A}$ such that $\beta$ is a single-path connected supporter to $\alpha$. The proof is analogous to the Case 1.

The proof for the REB and DF-QuAD semantics are analogous to the proof of the QE semantics.
\end{proof}

\begin{proposition}[Sign Invariance]
\label{prop_diff_polarity_relationship_sm}
For any $\alpha,\beta \in \mathcal{A}$, 
for any $h \in [-1,0) \cup (0,1]$,
if a semantics $\sigma$ satisfies directionality and monotonicity, then 
\begin{itemize}
    \item if $\beta$ is neutral to $\alpha$, $\Delta^{h}_{\beta \mapsto \alpha} = 0$;
    \item if $\beta$ is positive to $\alpha$, $\Delta^{h}_{\beta \mapsto \alpha} \geq 0$;
    \item if $\beta$ is negative to $\alpha$, $\Delta^{h}_{\beta \mapsto \alpha} \leq 0$.
\end{itemize}
\end{proposition}
\begin{proof}

1. Since $\beta$ is neutral to $\alpha$, $\sigma_{\tau'}(\alpha)=\sigma(\alpha)$, $\Delta^{h}_{\beta \mapsto \alpha}=\frac{\sigma_{\tau'}(\alpha)-\sigma(\alpha)}{h}=0$.

2. As proved in Proposition~\ref{prop_mono}, if an argument $\beta$ is positive to $\alpha$, then increasing $\tau(\beta)$ will not decrease $\sigma(\alpha)$. Hence, if $h>0$, $\sigma_{\tau'}(\alpha) \geq \sigma(\alpha)$; or if $h<0$, $\sigma_{\tau'}(\alpha) \leq \sigma(\alpha)$.
In either case, $\Delta^{h}_{\beta \mapsto \alpha}=\frac{\sigma_{\tau'}(\alpha)-\sigma(\alpha)}{h} \geq 0$.

3. Analogously to the proof in 2.

\end{proof}

\begin{proposition}[Alteration Existence]
Given 
a valid counterfactual $\tau^*$ (for the strong/$\delta-$approximate/weak counterfactual problem)
,
a topic argument $\alpha^* \in \mathcal{A}$ such that $\tau^*(\alpha^*)=\tau(\alpha^*)$,
and $\nexists \beta \in \mathcal{A}$ $(\beta \neq \alpha^*)$ such that $\beta$ is unknown to $\alpha^*$:
\begin{enumerate}
    \item If $\sigma(\alpha^*) \!\neq\! \sigma_{\tau^*}(\alpha^*)$, then 
    $\exists \!\gamma \!\in \!\mathcal{A}$ such that $\tau(\gamma) \!\neq\! \tau^*(\gamma)$;
    \item If $\sigma(\alpha^*) < \sigma_{\tau^*}(\alpha^*)$, then either $\exists \gamma \in \mathcal{A}$ such that $\gamma$ is positive to $\alpha^*$ and $\tau(\gamma) \leq \tau^*(\gamma)$ or $\exists \gamma \in \mathcal{A}$ such that $\gamma$ is negative to $\alpha^*$ and $\tau(\gamma) \geq \tau^*(\gamma)$;
    \item If $\sigma(\alpha^*) > \sigma_{\tau^*}(\alpha^*)$, then either $\exists \gamma \in \mathcal{A}$ such that $\gamma$ is positive to $\alpha^*$ and $\tau(\gamma) \geq \tau^*(\gamma)$ or $\exists \gamma \in \mathcal{A}$ such that $\gamma$ is negative to $\alpha^*$ and $\tau(\gamma) \leq \tau^*(\gamma)$.
\end{enumerate}
\end{proposition}
\begin{proof}

Since there is no unknown argument to $\alpha$, there are only neutral, positive, and negative arguments to $\alpha$. Since the final strength of $\alpha$ is different with the counterfactual $\tau^*$, we can exclude neutral arguments as they have no influence on $\alpha$ according to Definition~\ref{def_directionality}.
Therefore, we can only consider positive and negative arguments in the following proof.
As proved in Proposition~\ref{prop_mono}, if an argument $\beta$ is positive to $\alpha$, this means that increasing $\tau(\beta)$ will not decrease $\sigma(\alpha)$; if $\beta$ is negative to $\alpha$, then increasing $\tau(\beta)$ will not increase $\sigma(\alpha)$.

1. If the base scores for all the arguments are the same, then $\sigma(\alpha^*) \neq \sigma_{\tau^*}(\alpha^*)$ will not happen because $\sigma(\alpha^*) = \sigma_{\tau^*}(\alpha^*)$. Therefore, if $\sigma(\alpha^*) \neq \sigma_{\tau^*}(\alpha^*)$, then $\exists \gamma \in \mathcal{A}$ such that $\tau(\gamma) \neq \tau^*(\gamma)$.

2. If all the positive arguments decrease their base scores or all the negative arguments increase their base scores, then $\sigma(\alpha^*) < \sigma_{\tau^*}(\alpha^*)$ will not happen as there is no positive contribution to the final strength of $\alpha$. Therefore, either $\exists \gamma \in \mathcal{A}$ such that $\gamma$ is positive to $\alpha^*$ and $\tau(\gamma) \leq \tau^*(\gamma)$ or $\exists \gamma \in \mathcal{A}$ such that $\gamma$ is negative to $\alpha^*$ and $\tau(\gamma) \geq \tau^*(\gamma)$.

3. Analogously to the proof in 2.
\end{proof}

\begin{proposition}[Nullified-Validity]
Given 
a valid counterfactual $\tau^*$ (for the strong/$\delta-$approximate/weak counterfactual problem)
,
a topic argument $\alpha^* \in \mathcal{A}$ and a desired strength $s^*$ for $\alpha^*$, 
and another base score function $\tau^0$ such that 
for some
$\beta \in \mathcal{A}$ $(\beta \neq \alpha^*)$, $\tau^0(\beta)=0$ and $\tau^0(\gamma) = \tau^*(\gamma)$ for all $\gamma \in \mathcal{A} \setminus \{\beta\}$:
\begin{enumerate}
    \item If $\beta$ is neutral to $\alpha^*$,
    then $\tau^0$ is still 
    valid for the strong/$\delta-$approximate/weak counterfactual problem;
    \item If $\beta$ is negative to $\alpha^*$ and $\sigma_{\tau^*}(\alpha^*) \!\!\geq \!\! s^*$,
    then $\tau^0$ is still 
valid 
for the weak counterfactual problem;
    \item If $\beta$ is positive to $\alpha^*$ and $\sigma_{\tau^*}(\alpha^*) \leq s^*$,
    then $\tau^0$ is still 
    valid for the weak counterfactual problem.
\end{enumerate}
\end{proposition}
\begin{proof}

1. If $\beta$ is neutral to $\alpha^*$, then setting $\tau(\beta)$ to $0$ will not impact the final strength of $\alpha^*$, thus, $\tau^0$ is still valid for the weak counterfactual problems.

2. If the current strength is valid and greater or equal to the desired strength, then $\tau^0$ is still valid if it does not decrease the current strength.
Since $\beta$ is negative to $\alpha^*$, setting the base score of $\beta$ to $0$ will not decrease the final strength of $\alpha^*$.
Thus, $\tau^0$ is still valid for the weak counterfactual problem.

3. Analogously to the proof in 2.
\end{proof}

\begin{proposition}[Related-Validity]
Given 
a valid counterfactual explanation $\tau^*$ (for the weak counterfactual problem)
,
a topic argument $\alpha^* \in \mathcal{A}$ and a desired strength $s^*$ for $\alpha^*$ such that $\sigma_{\tau^*}(\alpha^*) \geq s^*$, for every other
base score function $\tau'$ in $\mathcal{Q}_{\tau'}$,
and all $\beta \in \mathcal{A}$ $(\beta \neq \alpha^*)$:
\begin{enumerate}
    \item If $\beta \in \mathcal{A}$ is neutral to $\alpha^*$, $\tau'(\beta) \geq \tau^*(\beta)$ and $\tau'(\gamma) = \tau^*(\gamma)$ for all 
    $\gamma \in \mathcal{A} \setminus \{\beta\}$, then $\tau'$ is also valid (for the weak counterfactual problem);
    \item If $\beta \in \mathcal{A}$ is negative to $\alpha^*$, $\tau'(\beta) \leq \tau^*(\beta)$ and $\tau'(\gamma) = \tau^*(\gamma)$ for all 
    $\gamma \in \mathcal{A} \setminus \{\beta\}$, then $\tau'$ is also a valid counterfactual (for the weak counterfactual problem);
    \item If $\beta \in \mathcal{A}$ is positive to $\alpha^*$, $\tau'(\beta) \geq \tau^*(\beta)$ and $\tau'(\gamma) = \tau^*(\gamma)$ for all 
    $\gamma \in \mathcal{A} \setminus \{\beta\}$, then $\tau'$ is also a valid counterfactual (for the weak counterfactual problem).
\end{enumerate}
\end{proposition}
\begin{proof}
1. If $\beta$ is neutral to $\alpha^*$, then increasing $\tau(\beta)$ will not impact the final strength of $\alpha^*$, thus, $\tau'$ is still valid for the weak counterfactual problems.

2. If the current strength is valid and greater or equal to the desired strength, then $\tau'$ is still valid if it does not decrease the current strength.
Since $\beta$ is negative to $\alpha^*$, decreasing the base score of $\beta$ will not decrease the final strength of $\alpha^*$.
Thus, $\tau'$ is still valid for the weak counterfactual problem.

3. Analogously to the proof in 2.
\end{proof}

\section{Additional Results for Section~\ref{sec_evaluation}}

\begin{table}[h]
\caption{Ablation studies for polarity and priority on \textbf{acyclic/cyclic} QBAFs under the REB gradual semantics: Comparison of average validity, $L_1$, $L_2$-norm distance, and runtime over 100 random generated acyclic full binary tree-like QBAFs with a depth of 7 and cyclic QBAFs with 100 arguments and 100 relations. The best results are shown in bold.}
\label{table_Expr_ablation_study_acyclic_reb}
    \centering
    \begin{tabular}{lcccc}
    \hline
        ~      & Validity  & $L_1$ & $L_2$ & Runtime (s)  \\ \hline
        BL     & 0.99/0.77 & 13.89/18.35 & 1.45/1.93 & 2.34/77.59  \\ 
        BL+pri & \textbf{1.00}/0.93 & \textbf{3.02}/\textbf{0.44} & \textbf{0.39}/\textbf{0.27} & 0.94/28.77  \\ 
        BL+pol & \textbf{1.00}/\textbf{1.00} & 22.32/1.50 & 2.08/0.48 & 0.02 /1.61 \\ 
        Ours   & \textbf{1.00}/\textbf{1.00} & 3.46/0.47 & 0.40/\textbf{0.27} & \textbf{0.01}/\textbf{0.69}  \\ \hline
    \end{tabular}
\end{table}

\begin{table}[h]
\caption{Ablation studies for polarity and priority on acyclic QBAFs under the DF-QuAD gradual semantics: Comparison of average validity, $L_1$, $L_2$-norm distance, and runtime over 100 random generated acyclic full binary tree-like QBAFs with a depth of 7. The best results are shown in bold.}
\label{table_Expr_ablation_study_acyclic_dfq}
    \centering
    \begin{tabular}{lcccc}
    \hline
        ~      & Validity  & $L_1$ & $L_2$ & Runtime (s)  \\ \hline
        BL     & \textbf{1.00} & 15.86 & 1.66 & 1.70 \\ 
        BL+pri & \textbf{1.00} & \textbf{6.89} & \textbf{0.85} & 2.67  \\ 
        BL+pol & \textbf{1.00} & 25.81 & 2.43 & \textbf{0.02} \\ 
        Ours   & \textbf{1.00} & 9.47 & 1.02 & \textbf{0.02}  \\ \hline
    \end{tabular}
\end{table}

\begin{figure}[h]
    \centering
    \includegraphics[width=0.75\columnwidth]{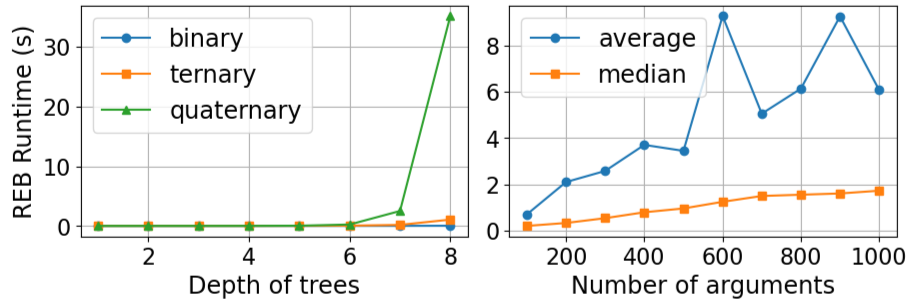}
    \caption{Scalability evaluation for CE-QArg on acyclic (left) and cyclic (right) QBAFs under \textbf{REB} gradual semantics: comparison of average runtime over 100 randomly generated acyclic and cyclic QBAFs.}
    \label{fig_scalability_acyclic_reb}
\end{figure}

\begin{figure}[h]
    \centering
    \includegraphics[width=0.4\columnwidth]{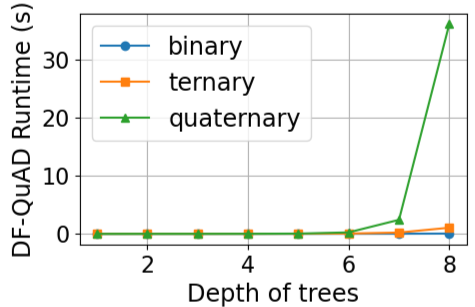}
    \caption{Scalability evaluation for CE-QArg on acyclic QBAFs under \textbf{DF-QuAD} gradual semantics: comparison of average runtime over 100 randomly generated acyclic QBAFs.}
    \label{fig_scalability_acyclic_dfq}
\end{figure}

\begin{figure}[h]
    \centering
    \includegraphics[width=0.9\columnwidth]{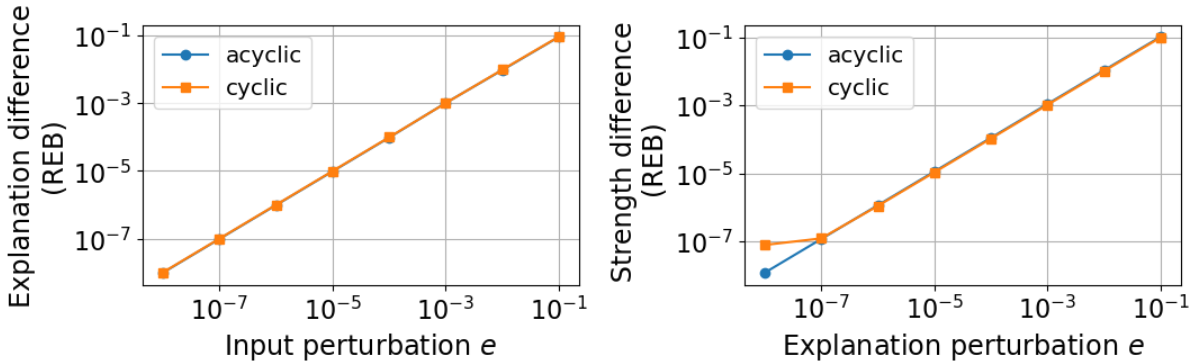}
    \caption{Robustness evaluation of perturbing base score functions (left) and perturbing generated counterfactuals (right) over 100 random generated QBAFs with increasing perturbation $e$ from $10^{-8}$ to $10^{-1}$ under \textbf{REB} gradual semantics. Acyclic QBAFs are binary trees with a depth of 7 while cyclic QBAFs contain 100 arguments and relations.}
    \label{fig_robustness_reb}
\end{figure}

\begin{figure}[h]
    \centering
    \includegraphics[width=0.9\columnwidth]{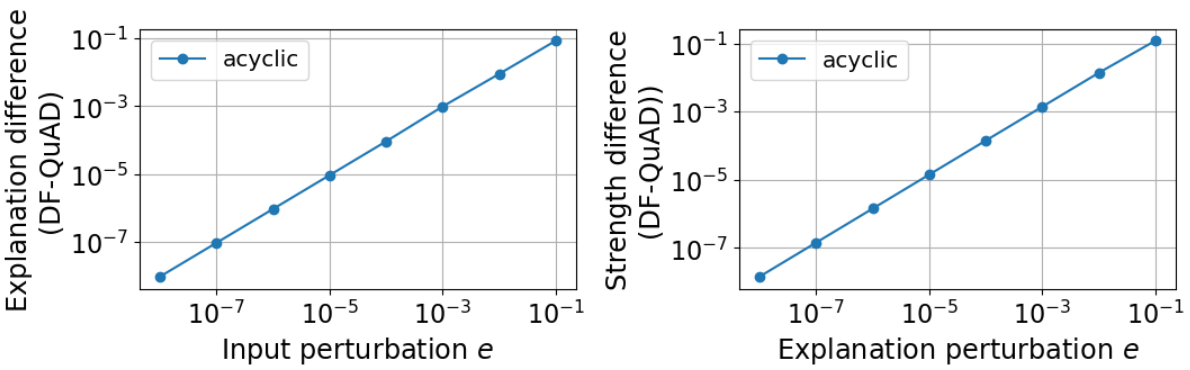}
    \caption{Robustness evaluation of perturbing base score functions (left) and perturbing generated counterfactuals (right) over 100 random generated QBAFs with increasing perturbation $e$ from $10^{-8}$ to $10^{-1}$ under \textbf{DF-QuAD} gradual semantics. Acyclic QBAFs are binary trees with a depth of 7.}
    \label{fig_robustness_dfq}
\end{figure}





\end{document}